%% file: neurips_2019.tex
\documentclass{article}

    \usepackage[preprint]{neurips_2019}

\input{math_commands.tex}

\usepackage[utf8]{inputenc} 
\usepackage[T1]{fontenc}    
\usepackage{hyperref}       
\usepackage{url}            
\usepackage{booktabs}       
\usepackage{amsfonts}       
\usepackage{nicefrac}       
\usepackage{microtype}      

\makeatletter
\g@addto@macro{\UrlBreaks}{\UrlOrds}
\makeatother
\usepackage{array}
\newcolumntype{L}{>{\centering\arraybackslash}m{9em}}
\usepackage{graphicx}
\usepackage{enumerate}
\usepackage{framed}
\usepackage[shortlabels]{enumitem}
\usepackage{xcolor}
\usepackage{svg}
\usepackage{algpseudocode}
\usepackage{algorithm}

\title{Supervised Learning on Relational Databases with Graph Neural Networks}

\author{Milan Cvitkovic \\
  Amazon Web Services \\
  \texttt{cvitkom@amazon.com} \\
}

\begin{document}

\maketitle

\begin{abstract}
The majority of data scientists and machine learning practitioners use relational data in their work [State of ML and Data Science 2017, Kaggle, Inc.].  But training machine learning models on data stored in relational databases requires significant data extraction and feature engineering efforts.  These efforts are not only costly, but they also destroy potentially important relational structure in the data.  We introduce a method that uses Graph Neural Networks to overcome these challenges.  Our proposed method outperforms state-of-the-art automatic feature engineering methods on two out of three datasets.
\end{abstract}

\section{Introduction}

Relational data is the most widely used type of data across all industries \citep{kaggle_inc_state_2017}.  Besides HTML/CSS/Javascript, relational databases (RDBs) are the most popular technology among developers \citep{stack_exchange_inc_stack_2018}.  The market merely for hosting RDBs is over \$45 billion USD \citep{asay_nosql_2016}, which is to say nothing of the societal value of the data they contain.

Yet learning on data in relational databases has received relatively little attention from the deep learning community recently. The standard strategy for working with RDBs in machine learning is to ``flatten'' the relational data they contain into tabular form, since most popular supervised learning methods expect their inputs to be fixed--size vectors.  This flattening process not only destroys potentially useful relational information present in the data, but the feature engineering required to flatten relational data is often the most arduous and time-consuming part of a machine learning practitioner's work. 

In what follows, we introduce a method based on Graph Neural Networks that operates on RDB data in its relational form without the need for manual feature engineering or flattening.  For two out of three supervised learning problems on RDBs, our experiments show that GNN-based methods outperform state-of-the-art automatic feature engineering methods.

\section{Background}
\subsection{Relational Databases}

\begin{figure}[h]
  \centering
  \includegraphics[width=\textwidth]{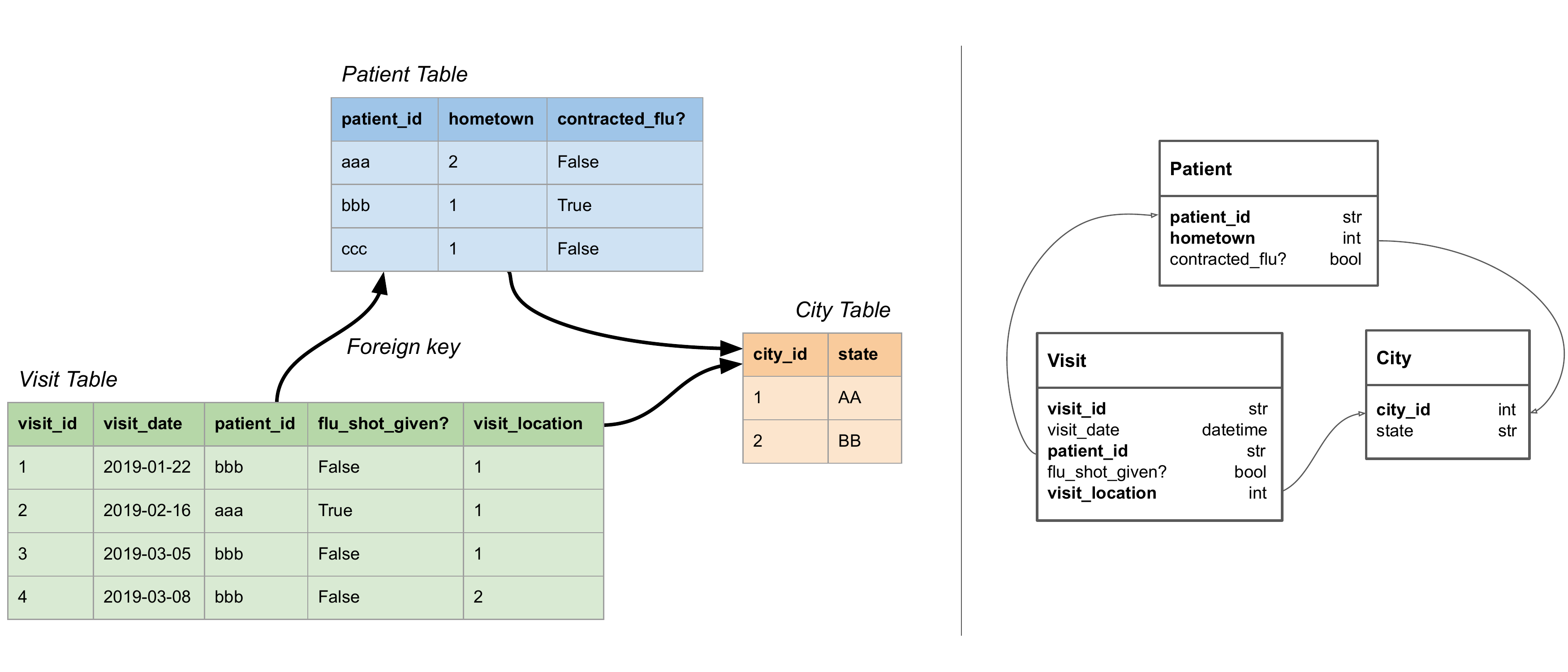}
  \caption{(Left) Toy example of a relational database (RDB). (Right) Schema diagram of the same RDB.}
\label{fig:RDBExample}
\end{figure}

A relational database\footnote{RDBs are sometimes informally called ``SQL databases''.} (RDB) is a set of tables, $\{\mT^1, \mT^2, \ldots, \mT^T\}$.  An example is shown in Figure \ref{fig:RDBExample}.   Each table represents a different type of entity: its rows correspond to different instances of that type of entity and its columns correspond to the features possessed by each entity.  Each column in a table may contain a different type of data, like integers, real numbers, text, images, date, times, geolocations, etc.   Unless otherwise specified, any entry in a table can potentially be empty, i.e. contain a \texttt{null} value.  We denote row $i$ of table $t$ as $\mT^t_{i, :}$ and column $j$ of table $t$ as $\mT^t_{:, j}$.

What makes RDBs ``relational'' is that the values in a column in one table can refer to rows of another table.  For example, in Figure \ref{fig:RDBExample} the column $\mT^\text{Visit}_{:, \text{patient\_id}}$   refers to rows in $\mT^\text{Patient}$ based on the values in $\mT^\text{Patient}_{:, \text{patient\_id}}$.  The value in $\mT^\text{Visit}_{i, \text{patient\_id}}$  indicates which patient came for Visit $i$.   A column like this that refers to another table is called a \emph{foreign key}.  

The specification of an RDB's tables, their columns, and the foreign key relationships between them is called the \emph{database schema}.  It is usually depicted diagrammatically, as on the right side of Figure \ref{fig:RDBExample}.

Readers familiar with object-oriented programming may find it helpful to think of each table as an object class.  In this analogy, the table's columns are the class's attributes, and each of the table's rows is an instance of that class.  A foreign key is an attribute that refers to an instance of a different class.

There are many software systems for creating and managing RDBs, including MySQL, PostgreSQL, and SQLite.  But effectively all RDB systems adhere closely to the same technical standard~\citep{international_organization_for_standardization_iso/iec_2016}, which defines how they are structured and what data types they can contain.  Thanks to this nonproliferation of standards, the ideas we present in this work apply to supervised learning problems in nearly every RDB in use today.

\subsection{Graph Neural Networks}

A Graph Neural Network (GNN) is any differentiable, parameterized function that takes a graph as input and produces an output by computing successively refined representations of the input.  For brevity, we defer explanation of how GNNs operate to Supplementary Material \ref{sec:suppGNNexplanation} and refer readers to useful surveys in \cite{gilmer_neural_2017}, \cite{battaglia_relational_2018}, and \cite{wu_comprehensive_2019}.

\section{Supervised learning on relational databases}\label{sec:supervisedlearning}

A broad class of learning problems on data in a relational database $D = \{\mT^1, \mT^2, \ldots, \mT^K\}$ can be formulated as follows: predict the values in a target column $\mT^k_{:, \text{target}}$ of $\mT^k \in D$ given all other relevant information in the database.\footnote{We ignore issues of target leakage for the purposes of this work, but in practice care is needed.}  In this work, \emph{supervised learning on relational databases} refers to this problem formulation.

This formulation encompasses all classification and regression problems in which we wish to predict the values in a particular column in $D$, or predict any aggregations or combinations of values in $D$.  This includes time series forecasting or predicting relationships between entities.  Traditional supervised learning problems like image classification or tabular regression are trivial cases of this formulation where $D$ contains one table.

There are several approaches for predicting values in $\mT^k_{:, \text{target}}$, including first-order logic- and graphical model inference-based approaches \citep{getoor_introduction_2007}.  In this work we consider the empirical risk minimization (ERM) approach.  The ERM approach is commonly used, though not always mentioned by name: it is the approach being implicitly used whenever a machine learning practitioner ``flattens'' or ``extracts'' or ``feature engineers'' data from an RDB into tabular form for use with a tabular, supervised, machine learning model.

More precisely, ERM assumes the entries of $\mT^k_{:, \text{target}}$ are sampled i.i.d. from some distribution, and we are interested in finding a function $f$ that minimizes the empirical risk
\begin{equation}\label{eq:ERM}
\min_{f \in \mathcal{F}} \frac{1}{N} \sum_{i=1}^N \mathcal{L}\left( f \left(\mT^1, \mT^2, \ldots, 
\mT^k_{:, : \setminus \text{target}}, \ldots, \mT^K \right), \mT^k_{i, \text{target}} \right)
\end{equation}
for a real-valued loss function $\mathcal{L}$ and a suitable function class $\mathcal{F}$, where $\mT^k_{:, : \setminus \text{target}}$ denotes table $k$ with the target column removed, and we assume that rows $1, \ldots, N$ of $\mT^k$ contain the training samples. 

To solve, or approximately solve, Equation \ref{eq:ERM}, we must choose a hypothesis class $\mathcal{F}$ and an optimization procedure over it.  We will accomplish this by framing supervised learning on RDBs in terms of learning tasks on graphs.

\subsection{Connection to learning problems on graphs}\label{sec:RDBasGraph}

A relational database can be interpreted as a directed graph --- specifically, a directed multigraph where each node has a type and has associated features that depend on its type.  The analogy is laid out in Table \ref{tab:RDBisGNN}.

\begin{table}[h]
\caption{Corresponding terms when interpreting a relational database as a graph.}
\label{tab:RDBisGNN}
\begin{tabular}{@{}ll@{}}
\toprule
\textbf{Relational Database} & \textbf{Graph} \\ \midrule
Row & Node \\
Table & Node type \\
Foreign key column & Edge type \\
Non-foreign-key column & Node feature \\
Foreign key reference from $\mT^A_{u,i}$ to $\mT^B_{v,j}$  & Directed edge from node $u$ of type $A$ to node $v$ of type $B$ \\
$i$th target value in table $k$, $\mT^k_{i, \text{target}}$ & Target feature on node $i$ of type $k$ \\
\bottomrule
\end{tabular}
\end{table}

Note that the RDB's schema diagram (an example of which is in Figure \ref{fig:RDBExample}) is not the same as this interpretation of the entire RDB as a graph. The former, which has tables as nodes, is a diagrammatic description of the properties of the latter, which has rows as nodes.

Note also that the RDB-as-graph interpretation is not bijective: directed multigraphs cannot in general be stored in RDBs by following the correspondence in Table \ref{tab:RDBisGNN}.\footnote{In principle one could use many-to-many tables to store any directed multigraph in an RDB, but this would not follow the correspondence in Table \ref{tab:RDBisGNN}.}  An RDB's schema places restrictions on which types of nodes can be connected by which types of edges, and on which features are associated with each node type, that general directed multigraphs may not satisfy.

The interpretation of an RDB as a graph shows that supervised learning on RDBs reduces to a node classification problem~\citep{atwood2016diffusion}. (For concision we only refer to classification problems, but our discussion applies equally to regression problems.)  In addition, the interpretation suggests that GNN methods are applicable to learning tasks on RDBs.

\subsection{Learning on RDBs with GNNs}

The first challenge in defining a hypothesis class $\mathcal{F}$ for use in Equation \ref{eq:ERM} is specifying how functions in $\mathcal{F}$ will interact with the RDB.  Equation \ref{eq:ERM} is written with $f \in \mathcal{F}$ taking the entire database $D$ (excluding the target values) as input.  But it is so written only for completeness.  In reality processing the entire RDB to produce a single output is intractable and inadvisable from a modeling perspective.  The algorithms in the hypothesis class $\mathcal{F}$ must retrieve and process only the information in the RDB that is relevant for making their prediction.

Stated in the language of node classification: we must choose how to select a subgraph of the full graph to use as input when predicting the label of a target node.  How best to do this in general in node classification is an active topic of research \citep{hamilton2017inductive}.  Models that can learn a strategy for selecting a subgraph from an RDB graph are an interesting prospect for future research.  In lieu of this, we present Algorithm \ref{alg:rdbtograph}, or \textproc{RDBToGraph}, a deterministic heuristic for selecting a subgraph.  \textproc{RDBToGraph} simply selects every ancestor of the target node, then selects every descendant of the target node and its ancestors.

\begin{algorithm}[h!]
\caption{\textproc{RDBToGraph}: Produce a graph of related entries in an RDB $D$ useful for classifying a target entry in $D$.  Runs in $O(|V| + |E|)$ time.}
\label{alg:rdbtograph}
\begin{algorithmic}
\Require RDB $D$ with target values removed, target row $i$, target table $k$. 
\Function{RDBToGraph}{$D$, $i$, $k$}\label{func:rdbtograph}
\State Let $G = (V, E)$ be $D$ encoded as a directed multigraph as per Table \ref{tab:RDBisGNN}.
\State Let $u \in V$ be the node corresponding to target entry $\mT^k_{i, \text{target}} \in D$
\State $V_S \gets \{ u \}$ 
\Repeat
    \State $A \gets \{ (v,w) \hspace{2pt} \vert \hspace{2pt} v \notin V_S \land w \in V_S \}$  // $A$ is for ancestors
    \State $V_S \gets V_S \cup \{ v \hspace{2pt} \vert \hspace{2pt} (v,w) \in A \} $
\Until {$V_S$ stops changing}
\Repeat
    \State $D \gets \{ (v,w) \hspace{2pt} \vert \hspace{2pt} v \in V_S \land w \notin V_S   \}$  // $D$ is for descendants
    \State $V_S \gets V_S \cup \{ w \hspace{2pt} \vert \hspace{2pt} (v,w) \in D \} $
\Until {$V_S$ stops changing}
\State $E_S \gets \{ (v,w) \hspace{2pt} \vert \hspace{2pt} v \in V_S \land w \in V_S \}$
\State \Return $(V_S, E_S)$
\EndFunction
\end{algorithmic}
\end{algorithm}

\textproc{RDBToGraph} is motivated by the ERM assumption and the semantics of RDBs.  ERM assumes all target nodes were sampled i.i.d.  Since the set of ancestors of a target node $u$ refer uniquely to $u$ through a chain of foreign keys, the ancestors of $u$ can be thought of as having been sampled along with $u$.  This is why \textproc{RDBToGraph} includes the ancestors of $u$ in the datapoint containing $u$.  The descendants of $u$ and its ancestors are included because, in the semantics of RDBs, a foreign key reference is effectively a type of feature, and we want to capture all potentially relevant features of $u$ (and its ancestors).

\textproc{RDBToGraph} misses potential gains that could be achieved by joining tables.  For example, in the KDD Cup 2014 dataset below, the Project table, which contains the target column, has a reverse foreign key to the Essay table.  In principle this means there could be multiple Essays for each Project.  But it so happens that there is one Essay for each Project in the dataset.  Knowing this, one could join the Project and Essay tables together into one table, reducing the number of nodes and edges in the graph datapoint.

\textproc{RDBToGraph} may also be a bad heuristic when a table in an RDB has foreign keys to itself.  Imagine a foreign key column in a table of employees that points to an employee's manager.  Applying \textproc{RDBToGraph} to make a prediction about someone at the bottom of the corporate hierarchy would result in selecting the entire database.  For such an RDB, we could modify \textproc{RDBToGraph} to avoid selecting too many nodes, for example by adding a restriction to follow each edge type only one time.  Nevertheless, \textproc{RDBToGraph} is a good starting heuristic and performs well for all datasets we present in this work.

\begin{figure}[h!]
  \centering
  \includegraphics[width=0.8\textwidth]{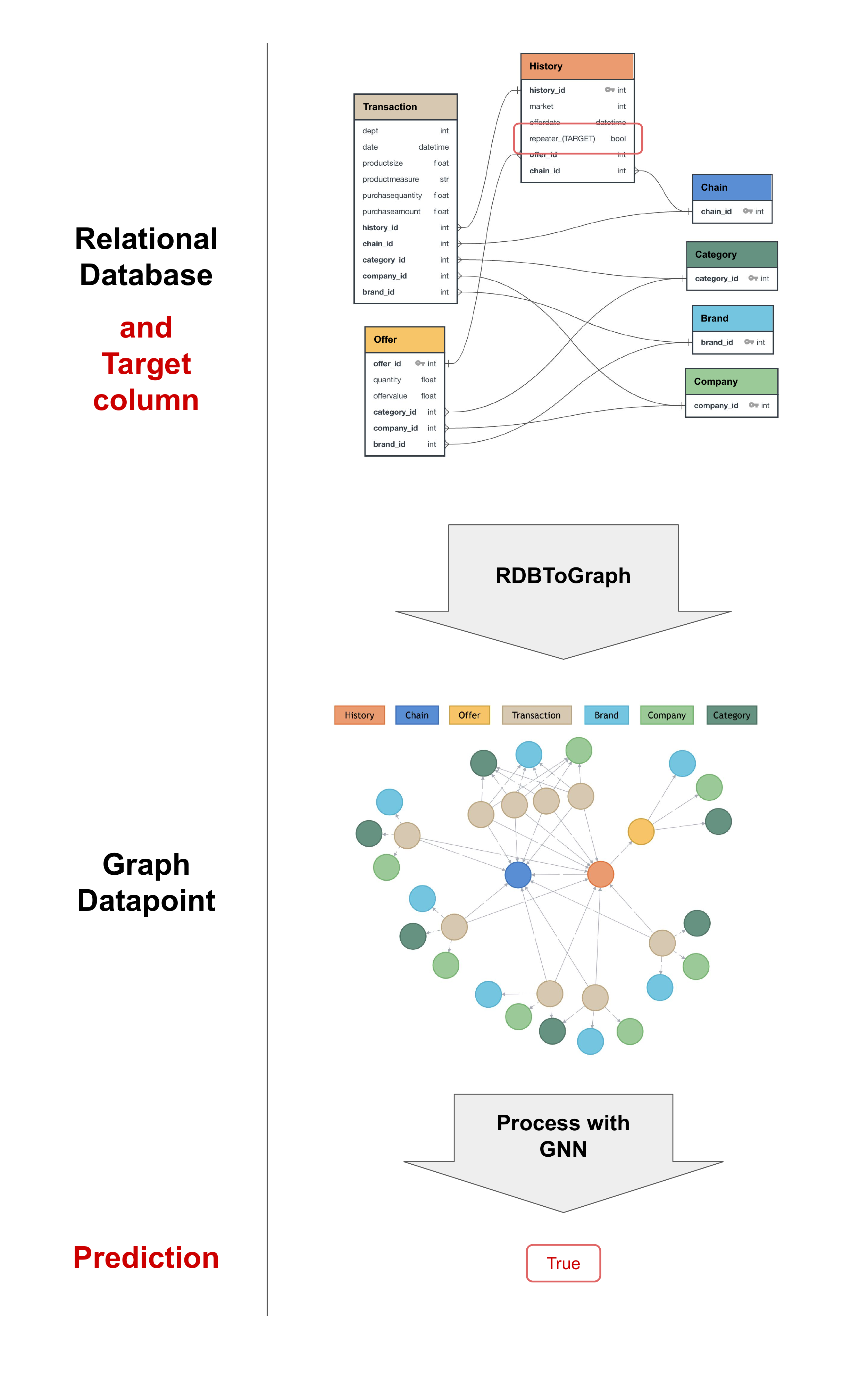}
  \caption{Example of our procedure for performing supervised learning on relational databases.  This example uses the Acquire Valued Shoppers Challenge RDB introduced in Section \ref{sec:datasets}. Given an RDB and a target column, we first use \textproc{RDBToGraph} to select the entries from the RDB that are relevant for predicting the target and assemble them into a directed multigraph.  Then we use a graph neural network to make a prediction of the target value.  Computing a differentiable loss function of the target and the output of $g_\theta$ lets us train $g_\theta$ by SGD.}
\label{fig:ModelExample}
\end{figure}

\textproc{RDBToGraph} followed by a GNN gives us a hypothesis class $\mathcal{F}$ suitable for optimizing Equation \ref{eq:ERM}.  In other words, for a GNN $g_\theta$ with parameters $\theta$, our optimization problem for performing supervised learning on relational databases is
\begin{equation}\label{eq:ERMRDBToCode}
\min_{\theta} \frac{1}{N} \sum_{i=1}^N \mathcal{L}\left( g_\theta \left( \textproc{RDBToGraph}\left( \left\{ \mT^1, \mT^2, \ldots, \mT^k_{:, : \setminus \text{target}}, \ldots, \mT^K \right\} , i, k \right) \right), \mT^k_{i, \text{target}}\right),
\end{equation} 
which we can perform using stochastic gradient descent.  Figure \ref{fig:ModelExample} shows an example of the entire procedure with the Acquire Valued Shoppers Challenge RDB introduced in Section \ref{sec:datasets}.

\section{Related Work}

The only work we are aware of that studies supervised learning tasks on relational databases of the sort considered above comes from the feature engineering literature. 

\citet{kanter_deep_2015} present a method called Deep Feature Synthesis (DFS) that automatically aggregates features from related tables to help predict the value in a user-selected target column in an RDB.  These engineered features can then be used as inputs to any standard tabular machine learning algorithm.  DFS performs feature aggregation by recursively applying functions like \texttt{MAX} or \texttt{SUM} to rows connected to the target column by foreign key relationships.  This is somewhat analogous to the multi-view learning approach of \cite{guo_multirelational_2008}.  

The main advantages of DFS over a GNN-based approach like the one we are proposing are (1) it produces more interpretable features, and (2) by converting RDB data into tabular features, it lets the user apply powerful tabular learning methods like gradient boosted decision trees \citep{ke2017lightgbm} to their problem, which have been observed to generally perform better on tabular data than existing deep models \citep{arik2019tabnet}.  The disadvantages include the combinatorial explosion of possible aggregated features that must searched over and the lack of ability to learn these feature end-to-end.  These disadvantages mean users must either be judicious about how many features they wish to engineer or pay a large computational cost.  In addition, a system based on DFS may miss out on potential gains from transfer learning that are possible with deep methods like GNNs. 

\citet{lam_one_2017} and \citet{lam_neural_2018} both extend \cite{kanter_deep_2015}, the former by expanding the types of feature quantizations and embeddings used, and the latter by using Recurrent Neural Networks as the aggregation functions rather than functions like \texttt{MAX} and \texttt{SUM}.

A more loosely related area of prior work is Statistical Relational Learning \citep{getoor_introduction_2007}, specifically Probabilistic Relational Models  (PRMs) \citep{koller2007introduction}.  PRMs define a joint probability distribution over the entities in an RDB schema.  Learning a PRM amounts to estimating parameters for this joint model from the contents of a particular RDB.  Supervised learning on RDBs is somewhat similar to learning in PRMs, except we are only interested in learning one particular conditional distribution (that of the target conditional on other entries in the RDB) rather than learning a joint model of the all entities in the RDB schema.`

Work on modeling heterogeneous networks \citep{shi2016survey} is also relevant.  Heterogeneous networks are examples of directed multigraphs, and thus techniques applicable to modeling them, including recent GNN-based techniques \citep{wang2019heterogeneous}, may prove useful for supervised learning on RDBs, though we do not explore them in this work.

\section{Datasets}\label{sec:datasets}

Despite the ubiquity of supervised learning problems on data from relational databases, there are few public RDB datasets available to the machine learning research community.   This is a barrier to research into this important area.

As part of this work, we provide code for converting data from three public Kaggle\footnote{\url{http://www.kaggle.com/}} competitions into RDB format.  The datasets are the Acquire Valued Shoppers Challenge, the KDD Cup 2014, and the Home Credit Default Risk Challenge.  All are binary classification tasks with competition performance measured by the area under the receiver operating characteristic curve (AUROC).
 
Basic information about the datasets is given in Table \ref{tab:dsinfo}.  Detailed information about each dataset is given in Appendix \ref{sec:suppdatasetinfo}.  

\begin{table}[h]
\caption{Dataset summary information}
\label{tab:dsinfo}
\begin{tabular}{@{}rLLL@{}}
\toprule
 & Acquire Valued Shoppers Challenge & Home Credit Default Risk & KDD Cup 2014 \\ \midrule
\multicolumn{1}{l|}{Train datapoints } 
& 160,057 & 307,511 & 619,326 \\
\multicolumn{1}{l|}{Tables/Node types} 
& 7 & 7 & 4 \\
\multicolumn{1}{l|}{Foreign keys/Edge types} 
& 10 & 9 & 4 \\
\multicolumn{1}{l|}{Feature types}
& Categorical, Scalar, Datetime & Categorical, Scalar & Categorical, Geospatial, Scalar, Textual, Datetime \\
\midrule
\multicolumn{4}{c}{ \url{www.kaggle.com/c/acquire-valued-shoppers-challenge} } \\
\multicolumn{4}{c}{ \url{www.kaggle.com/c/home-credit-default-risk/overview/evaluation} } \\
\multicolumn{4}{c}{ \url{www.kaggle.com/c/kdd-cup-2014-predicting-excitement-at-donors-choose} } \\
\bottomrule
\end{tabular}
\end{table}

\section{Experiments}
Code to reproduce all experiments may be found online.\footnote{\url{https://github.com/mwcvitkovic/Supervised-Learning-on-Relational-Databases-with-GNNs}}

We compare the performance of GNN-based methods of the type proposed in Section \ref{sec:supervisedlearning} to state-of-the-art tabular models and feature engineering approaches.  Table \ref{tab:relative-auroc} shows the AUROC performance of models relative to the best-performing tabular model on each dataset.  We compare relative performance rather than absolute performance since for some datasets the variance in performance between cross-validation splits is larger than the variance of performance between algorithms.  Supplementary Tables \ref{tab:absolute-auroc} and \ref{tab:absolute-acc} give the absolute AUROC and accuracy results for the interested reader.

The tabular models we use in our baselines are logistic regression (LogReg), a 2-hidden-layer multilayer perceptron (MLP), and the \texttt{LightGBM}\footnote{\url{https://lightgbm.readthedocs.io}} gradient boosted tree algorithm \citep{ke2017lightgbm} (GBDT).   ``Single-table'' means the tabular model was trained only on data from the RDB table that contains the target column, ignoring the other tables in the RDB.  ``DFS'' means that the Deep Feature Synthesis method of \citet{kanter_deep_2015}, as implemented in the \texttt{featuretools}\footnote{\url{https://docs.featuretools.com/}} library, was used to engineer features from all the tables in the RDB for use by the tabular model.

The standard GNN models we use as the learned part $g_\theta$ of Equation \ref{eq:ERMRDBToCode} are the Graph Convolutional Network of \citet{kipf2016semi} (GCN), the Graph Isomorphism Network of \citet{ginpaper} (GIN), and the Graph Attention Network of \citet{gatpaper} (GAT).  Given that each table in an RDB contains different features and each foreign key encodes a different type of relationship, we also test versions of these standard GNNs which maintain different parameters for each node-update and message-passing function, depending on the node type and edge type respectively.  These are in the spirit of \cite{Schlichtkrull2017ModelingRD} and \cite{wang2019heterogeneous}, though are different models.  We refer to these as ERGCN, ERGIN, and ERGAT, respectively (ER for ``entity-relational'').   Additionally, inspired by \cite{Luzhnica2019OnGC}, we compare against a baseline called PoolMLP, which does no message passing and computes the output merely by taking the mean of all node hidden states ($\vh_v^0$ in the notation of Supplementary Material \ref{sec:suppGNNexplanation}) and passing this through a 1-hidden-layer MLP.

Finally, since we noticed evidence of overfitting in the GNN models during training (even with significant regularization), we tested whether a stacking strategy \citep{wolpert1992stacked} that trains a GBDT on a concatenation of single-table features and the pre-logit activations of a trained GNN improves performance.  We denote this as ``Stacked''.

Thorough details about model and experiment implementation are in Appendix \ref{sec:implementationdetails}.

\begin{table}[h]
\caption{Performance of baseline (above the line) and our GNN-based (below the line) learning algorithms on three supervised learning problems on RDBs.  Values are the AUROC metric relative to the single-table logistic regression baseline; they are reported as the mean over 5 cross-validation splits, plus or minus the standard deviation.  Bold values are those within one standard deviation of the maximum in the column. Larger values are better.}
\label{tab:relative-auroc}  
\begin{tabular}{@{}rLcc@{}}
\toprule
 & Acquire Valued Shoppers Challenge & Home Credit Default Risk & KDD Cup 2014 \\ \midrule
\multicolumn{1}{l|}{Single-table LogReg} 
&  $ 0 \pm 0 $ &                      $ 0 \pm 0 $ &                      $ 0 \pm 0 $ \\
\multicolumn{1}{l|}{Single-table MLP}
&  $ -0.0007 \pm 0.0009 $ &            $ 0.0021 \pm 0.0007 $ &              $ 0.014 \pm 0.002 $ \\
\multicolumn{1}{l|}{Single-table GBDT}
&  $ 0.0043 \pm 0.0008 $ &              $ 0.006 \pm 0.003 $ &              $ \mathbf{0.027 \pm 0.001} $ \\
\multicolumn{1}{l|}{DFS + LogReg} 
&  $ 0.0106 \pm 0.0007 $ &               $ 0.023 \pm 0.002 $ &   $ 0.004 \pm 0.002 $ \\
\multicolumn{1}{l|}{DFS + MLP}
&  $ 0.0087 \pm 0.0009 $ &               $ 0.016 \pm 0.003 $ &   $ 0.007 \pm 0.001 $ \\
\multicolumn{1}{l|}{DFS + GBDT}
&  $ 0.003 \pm 0.001 $ &               $ 0.029 \pm 0.002 $ &   $ \mathbf{0.027 \pm 0.002} $ \\
\multicolumn{1}{l|}{PoolMLP}
&  $ 0.006 \pm 0.002 $ &   $ 0.021 \pm 0.003 $ &  $ 0.007 \pm 0.003 $ \\
\multicolumn{1}{l|}{Stacked PoolMLP}
&  $ 0.008 \pm 0.002 $ &   $ 0.023 \pm 0.003 $ &  $ 0.016 \pm 0.004 $ \\
\midrule
\multicolumn{1}{l|}{GCN}
&  $ \mathbf{0.038 \pm 0.002} $ &   $ \mathbf{0.032 \pm 0.002} $ &  $ 0.013 \pm 0.002 $ \\
\multicolumn{1}{l|}{GIN}
&  $ 0.035 \pm 0.005 $ &   $ 0.027 \pm 0.002 $ &  $ 0.014 \pm 0.001 $ \\
\multicolumn{1}{l|}{GAT}
&  $ 0.032 \pm 0.003 $ &   $ \mathbf{0.031 \pm 0.002} $ &  $ 0.013 \pm 0.002 $ \\
\multicolumn{1}{l|}{ERGCN}
&  $ \mathbf{0.040 \pm 0.002} $ &   $ \mathbf{0.030 \pm 0.002} $ &  $ 0.013 \pm 0.002 $ \\
\multicolumn{1}{l|}{ERGIN}
&  $ \mathbf{0.039 \pm 0.002} $ &   $ 0.025 \pm 0.003 $ &  $ 0.015 \pm 0.002 $ \\
\multicolumn{1}{l|}{ERGAT}
&  $ \mathbf{0.039 \pm 0.001} $ &   $ \mathbf{0.031 \pm 0.002} $ &  $ 0.015 \pm 0.001 $ \\
\multicolumn{1}{l|}{Stacked GCN}
&  $ 0.037 \pm 0.002 $ &   $ \mathbf{0.030 \pm 0.002} $ &  $ 0.008 \pm 0.004 $ \\
\multicolumn{1}{l|}{Stacked GIN}
&  $ 0.033 \pm 0.006 $ &   $ 0.027 \pm 0.002 $ &  $ 0.010 \pm 0.006 $ \\
\multicolumn{1}{l|}{Stacked GAT}
&  $ 0.036 \pm 0.002 $ &   $ \mathbf{0.032 \pm 0.002} $ &  $ 0.012 \pm 0.002 $ \\
\multicolumn{1}{l|}{Stacked ERGCN}
&  $ \mathbf{0.038 \pm 0.003} $ &   $ 0.029 \pm 0.002 $ &  $ 0.013 \pm 0.005 $ \\
\multicolumn{1}{l|}{Stacked ERGIN}
&  $ \mathbf{0.039 \pm 0.002} $ &   $ 0.027 \pm 0.002 $ &  $ 0.015 \pm 0.005 $ \\
\multicolumn{1}{l|}{Stacked ERGAT}
&  $ 0.0379 \pm 0.0002 $ &   $ \mathbf{0.030 \pm 0.002} $ &  $ 0.010 \pm 0.005 $ \\
\bottomrule
\end{tabular}
\end{table}

\section{Discussion}
The results in Table \ref{tab:relative-auroc} suggest that GNN-based methods are a valuable new approach for supervised learning on RDBs.

GNN-based methods perform significantly better than the best baseline method on the Acquire Valued Shoppers Challenge dataset and perform moderately better than the best baseline on the Home Credit Default Risk dataset.  They perform worse than the best baseline on the KDD Cup 2014, however no feature engineering approach of any kind offers an advantage on that dataset.  The determinant of success on the KDD Cup 2014 dataset does not seem to be information outside the dataset's main table.

Interestingly, as one can see from the RDB schema diagrams of the datasets in Supplementary Section \ref{sec:suppdatasetinfo}, the more tables and foreign key relationships an RDB has, the better GNN-based methods perform on it, with the KDD Cup 2014 dataset being the least ``relational'' of the datasets and the one on which GNNs seem to offer no benefit.

Though every GNN-based method we tested matches or exceeds the best baseline method on the Acquire Valued Shoppers Challenge and Home Credit Default Risk datasets, there is no clearly superior GNN model among them.  Neither the ER models nor stacking offers a noticeable benefit.  But we emphasize that all our experiments used off-the-shell GNNs or straightforward modifications thereof --- the space of possible RDB-specific GNNs is large and mostly unexplored.

We do not explore it in this work, but we suspect the greatest advantage of using GNN-based models for supervised learning on RDBs, as opposed to other feature engineering approaches, is the potential to leverage transfer learning.  GNN models can be straightforwardly combined with, e.g., pretrained Transformer representations of \texttt{TEXT} inputs in the KDD Cup 2014 dataset, and trained end-to-end.

\subsubsection*{Acknowledgments}

We thank Da Zheng, and Minjie Wang, and Zohar Karnin for helpful discussions.

\bibliography{references}
\bibliographystyle{plainnat}

\appendix

\section{Graph Neural Networks}\label{sec:suppGNNexplanation}

Many types of GNN have been introduced in the literature, and several nomenclatures and taxonomies have been proposed.  This is a recapitulation of the Message Passing Neural Network formalism of GNNs from \cite{gilmer_neural_2017}.  Most GNNs can be described in this framework, though not all, such as \cite{murphy2019relational}.

A (Message Passing) GNN $f$ takes as input a graph $G = (V,E)$.  Each vertex $v \in V$ has associated features $\vx_v$, and each edge $(v,w) \in E$ may have associated features $\ve_{vw}$.  The output of the GNN, $f(G)$, is computed by the following steps:

\begin{enumerate}
\item A function $S$ is used to initialize a hidden state vector $\vh_v^0 \in \mathbb{R}^d$ for each vertex:
\[
\vh_v^0 = S(\vx_v), \ \ \forall v \in V.
\]
E.g. if $\vx_v$ is a sentence, $S$ could be a tf-idf transform.
\item For each iteration $t$ from 1 to $T$:
    \begin{enumerate}
    \item Each vertex $v$ sends a ``message''
    \[
    \vm_{vw}^{t} = M_t(\vh_v^{t-1}, \vh_w^{t-1}, \ve_{vw}), \ \ \forall w \in \mathcal{N}_v
    \]
    to each of its neighbors $w$, where $\mathcal{N}_v$ is the set of all neighbors of $v$.
    \item Each vertex $v$ aggregates the messages it received using a function $A_t$, where $A_t$ takes a variable number of arguments and is invariant to permutations of its arguments:
    \[
    \vm_v^{t} = A_t(\{\vm_{vw}^{t} \vert w \in \mathcal{N}_v \}).
    \]
    \item Each vertex $v$ updates its hidden state as a function of its current hidden state and the aggregated messages it received:
    \[
    \vh_v^{t} = U_t(\vh_{v}^{t-1}, \vm_v^{t}).
    \]
    \end{enumerate}
\item The output $f(G)$ is computed via the ``readout'' function $R$, which is invariant to the permutation of its arguments:
\[
f(G) = R(\{ \vh_v^{T} \vert v \in V \}).
\]
\end{enumerate}
    
The functions $R$, $U_t$, $A_t$, $M_t$, and optionally $S$ are differentiable, possibly parameterized functions, so the GNN may be trained by using stochastic gradient descent to minimize a differentiable loss function of $f(G)$.

\section{Additional Results}

\begin{table}[h]
\caption{AUROC of baseline (above the line) and our GNN-based (below the line) learning algorithms on three supervised learning problems on RDBs.  Values are the mean over 5 cross-validation splits, plus or minus the standard deviation.  Larger values are better.}
\label{tab:absolute-auroc}
\begin{tabular}{@{}rLcc@{}}
\toprule
 & Acquire Valued Shoppers Challenge & Home Credit Default Risk & KDD Cup 2014 \\ \midrule
\multicolumn{1}{l|}{Single-table LogReg} 
&  $ 0.686 \pm 0.002 $ &              $ 0.748 \pm 0.004 $ &              $ 0.774 \pm 0.002 $ \\
\multicolumn{1}{l|}{Single-table MLP}
&  $ 0.685 \pm 0.002 $ &              $ 0.750 \pm 0.004 $ &              $ 0.788 \pm 0.002 $ \\
\multicolumn{1}{l|}{Single-table GBDT}
&  $ 0.690 \pm 0.002 $ &              $ 0.754 \pm 0.004 $ &              $ 0.801 \pm 0.002 $ \\
\multicolumn{1}{l|}{DFS + LogReg} 
&  $ 0.696 \pm 0.001 $ &               $ 0.771 \pm 0.005 $ &   $ 0.778 \pm 0.002 $ \\
\multicolumn{1}{l|}{DFS + MLP}
&  $ 0.694 \pm 0.001 $ &               $ 0.764 \pm 0.005 $ &   $ 0.781 \pm 0.002 $ \\
\multicolumn{1}{l|}{DFS + GBDT}
&  $ 0.689 \pm 0.003 $ &               $ 0.777 \pm 0.004 $ &   $ 0.801 \pm 0.003 $ \\
\multicolumn{1}{l|}{PoolMLP}
&  $ 0.692 \pm 0.001 $ &   $ 0.769 \pm 0.005 $ &  $ 0.781 \pm 0.003 $ \\
\multicolumn{1}{l|}{Stacked PoolMLP}
&  $ 0.694 \pm 0.003 $ &   $ 0.771 \pm 0.007 $ &  $ 0.790 \pm 0.005 $ \\
\midrule
\multicolumn{1}{l|}{GCN}
&  $ 0.723 \pm 0.002 $ &   $ 0.780 \pm 0.004 $ &  $ 0.787 \pm 0.003 $ \\
\multicolumn{1}{l|}{GIN}
&  $ 0.721 \pm 0.006 $ &   $ 0.775 \pm 0.005 $ &  $ 0.788 \pm 0.002 $ \\
\multicolumn{1}{l|}{GAT}
&  $ 0.717 \pm 0.002 $ &   $ 0.778 \pm 0.005 $ &  $ 0.787 \pm 0.002 $ \\
\multicolumn{1}{l|}{ERGCN}
&  $ 0.726 \pm 0.002 $ &   $ 0.778 \pm 0.005 $ &  $ 0.787 \pm 0.003 $ \\
\multicolumn{1}{l|}{ERGIN}
&  $ 0.725 \pm 0.003 $ &   $ 0.773 \pm 0.005 $ &  $ 0.788 \pm 0.002 $ \\
\multicolumn{1}{l|}{ERGAT}
&  $ 0.725 \pm 0.001 $ &   $ 0.779 \pm 0.005 $ &  $ 0.789 \pm 0.002 $ \\
\multicolumn{1}{l|}{Stacked GCN}
&  $ 0.722 \pm 0.002 $ &   $ 0.778 \pm 0.003 $ &  $ 0.782 \pm 0.003 $ \\
\multicolumn{1}{l|}{Stacked GIN}
&  $ 0.719 \pm 0.007 $ &   $ 0.775 \pm 0.005 $ &  $ 0.784 \pm 0.006 $ \\
\multicolumn{1}{l|}{Stacked GAT}
&  $ 0.722 \pm 0.001 $ &   $ 0.779 \pm 0.005 $ &  $ 0.786 \pm 0.003 $ \\
\multicolumn{1}{l|}{Stacked ERGCN}
&  $ 0.724 \pm 0.003 $ &   $ 0.777 \pm 0.005 $ &  $ 0.786 \pm 0.005 $ \\
\multicolumn{1}{l|}{Stacked ERGIN}
&  $ 0.725 \pm 0.003 $ &   $ 0.775 \pm 0.006 $ &  $ 0.789 \pm 0.006 $ \\
\multicolumn{1}{l|}{Stacked ERGAT}
&  $ 0.724 \pm 0.002 $ &   $ 0.778 \pm 0.005 $ &  $ 0.784 \pm 0.006 $ \\
\bottomrule
\end{tabular}
\end{table}

\begin{table}[h]
\caption{Percent accuracy of baseline (above the line) and our GNN-based (below the line) learning algorithms on three supervised learning problems on RDBs.  Values are the mean over 5 cross-validation splits, plus or minus the standard deviation.  Larger values are better.}
\label{tab:absolute-acc}
\begin{tabular}{@{}rLcc@{}}
\toprule
 & Acquire Valued Shoppers Challenge & Home Credit Default Risk & KDD Cup 2014 \\ \midrule
\multicolumn{1}{l|}{Guess Majority Class}
& $72.9$ & $ 91.9 $ & $94.07$ \\
\multicolumn{1}{l|}{Single-table LogReg} 
&  $ 73.1 \pm 0.2 $ &                 $ 91.9 \pm 0.1 $ &               $ 94.07 \pm 0.07 $ \\
\multicolumn{1}{l|}{Single-table MLP}
&  $ 73.2 \pm 0.2 $ &                 $ 91.9 \pm 0.1 $ &               $ 94.07 \pm 0.07 $ \\
\multicolumn{1}{l|}{Single-table GBDT}
&  $ 73.3 \pm 0.2 $ &                 $ 91.9 \pm 0.1 $ &               $ 94.06 \pm 0.07 $ \\
\multicolumn{1}{l|}{DFS + LogReg} 
&  $ 73.5 \pm 0.2 $ &                  $ 91.9 \pm 0.1 $ &    $ 94.07 \pm 0.07 $ \\
\multicolumn{1}{l|}{DFS + MLP}
&  $ 73.6 \pm 0.3 $ &                  $ 91.9 \pm 0.1 $ &    $ 94.07 \pm 0.07 $ \\
\multicolumn{1}{l|}{DFS + GBDT}
&  $ 73.4 \pm 0.3 $ &                $ 91.96 \pm 0.09 $ &    $ 94.06 \pm 0.07 $ \\
\multicolumn{1}{l|}{PoolMLP}
&  $ 73.6 \pm 0.1 $ &      $ 91.9 \pm 0.1 $ &  $ 94.07 \pm 0.07 $ \\
\multicolumn{1}{l|}{Stacked PoolMLP}
&  $ 73.6 \pm 0.2 $ &      $ 91.9 \pm 0.1 $ &  $ 94.04 \pm 0.08 $ \\
\midrule
\multicolumn{1}{l|}{GCN}
&  $ 75.4 \pm 0.1 $ &    $ 91.96 \pm 0.09 $ &  $ 94.07 \pm 0.07 $ \\
\multicolumn{1}{l|}{GIN}
&  $ 75.4 \pm 0.3 $ &      $ 91.9 \pm 0.1 $ &  $ 94.07 \pm 0.07 $ \\
\multicolumn{1}{l|}{GAT}
&  $ 72.9 \pm 0.3 $ &      $ 91.1 \pm 0.3 $ &  $ 94.07 \pm 0.07 $ \\
\multicolumn{1}{l|}{ERGCN}
&  $ 75.6 \pm 0.2 $ &      $ 91.9 \pm 0.1 $ &  $ 94.07 \pm 0.07 $ \\
\multicolumn{1}{l|}{ERGIN}
&  $ 75.6 \pm 0.3 $ &      $ 91.9 \pm 0.1 $ &  $ 94.07 \pm 0.07 $ \\
\multicolumn{1}{l|}{ERGAT}
&  $ 75.2 \pm 0.2 $ &      $ 92.0 \pm 0.1 $ &  $ 94.07 \pm 0.07 $ \\
\multicolumn{1}{l|}{Stacked GCN}
&  $ 75.5 \pm 0.1 $ &      $ 91.9 \pm 0.1 $ &  $ 94.06 \pm 0.07 $ \\
\multicolumn{1}{l|}{Stacked GIN}
&  $ 75.4 \pm 0.3 $ &      $ 92.0 \pm 0.1 $ &  $ 94.04 \pm 0.06 $ \\
\multicolumn{1}{l|}{Stacked GAT}
&  $ 75.5 \pm 0.3 $ &      $ 92.0 \pm 0.1 $ &  $ 94.07 \pm 0.07 $ \\
\multicolumn{1}{l|}{Stacked ERGCN}
&  $ 75.6 \pm 0.3 $ &      $ 91.9 \pm 0.1 $ &  $ 94.02 \pm 0.06 $ \\
\multicolumn{1}{l|}{Stacked ERGIN}
&  $ 75.6 \pm 0.2 $ &      $ 91.9 \pm 0.1 $ &  $ 94.04 \pm 0.07 $ \\
\multicolumn{1}{l|}{Stacked ERGAT}
&  $ 75.6 \pm 0.2 $ &      $ 91.9 \pm 0.1 $ &  $ 94.06 \pm 0.07 $ \\
\bottomrule
\end{tabular}
\end{table}

\section{Experiment Details}\label{sec:implementationdetails}

\subsection{Software and Hardware}

The \texttt{LightGBM}\footnote{\url{https://lightgbm.readthedocs.io}} library \citep{ke2017lightgbm} was used to implement the GBDT models. 
All other models were implemented using the \texttt{PyTorch}\footnote{\url{https://pytorch.org/}} library \citep{pytorchpaper}.  GNNs were implemented using the \texttt{DGL}\footnote{\url{https://www.dgl.ai/}} library \citep{dglpaper}.  
All experiments were run on an Ubuntu Linux machine with 8 CPUs and 60GB memory, with all models except for the GBDTs trained using a single NVIDIA V100 Tensor Core GPU.  Creating the DFS features was done on an Ubuntu Linux machine with 48 CPUs and 185GB memory.

\subsection{GNN implementation}

Adopting the nomenclature from Supplementary Material \ref{sec:suppGNNexplanation}, once the input graph has been assembled by \texttt{RDBToGraph}, the initialization function $S$ for converting the features $\vx_v$ of each vertex $v$ into a real--valued vector in $\mathbb{R}^{d}$ proceeded as follows: (1) each of the vertex's features was vectorized according to the feature's data type, (2) these vectors were concatenated, (3) the concatenated vector was passed through an single-hidden-layer MLP with output dimension $\mathbb{R}^{d}$.  Each node type uses its own single-hidden-layer MLP initializer.  The dimension of the hidden layer was 4x the dimension of the concatenated features.

To vectorize columns containing scalar data, $S$ normalizes them to zero median and unit interquartile range\footnote{\url{https://scikit-learn.org/stable/modules/generated/sklearn.preprocessing.RobustScaler.html}} and appends a binary flag for missing values.  To vectorize columns containing categorical information, $S$ uses a trainable embedding with dimension the minimum of 32 or the cardinality of categorical variable.  To vectorize \texttt{text} columns, $S$ simply encodes the number of words in the text and the length of the text.  To vectorize \texttt{latlong} columns, $S$ concatenates the following:
\begin{enumerate}
    \item $\cos(lat) * \cos(long)$
    \item $y = \cos(lat) * \sin(long)$
    \item $z = \sin(lat)$
    \item $lat / 90$
    \item $long / 180$
\end{enumerate}
And to vectorize \texttt{datetime} columns, $S$ concatenates the following commonly used date and time features:
\begin{enumerate}
    \item Year (scalar value)
    \item Month (one-hot encoded)
    \item Week (one-hot encoded)
    \item Day (one-hot encoded)
    \item Day of week (one-hot encoded)
    \item Day of year (scalar value)
    \item Month end? (bool, one-hot encoded)
    \item Month start? (bool, one-hot encoded)
    \item Quarter end? (bool, one-hot encoded)
    \item Quarter start? (bool, one-hot encoded)
    \item Year end? (bool, one-hot encoded)
    \item Year start? (bool, one-hot encoded)
    \item Day of week $\cos$ (scalar value)
    \item Day of week $\sin$ (scalar value)
    \item Day of month $\cos$ (scalar value)
    \item Day of month $\sin$ (scalar value)
    \item Month of year $\cos$ (scalar value)
    \item Month year $\sin$ (scalar value)
    \item Day of year $\cos$ (scalar value)
    \item Day of year $\sin$ (scalar value)
\end{enumerate}

The $\cos$ and $\sin$ values are for representing cyclic information, and are given by computing $\cos$ or $\sin$ of $2 \pi \frac{value}{period}$.  E.g. ``day of week $\cos$'' for Wednesday, the third day of seven in the week, is $\cos(2 \pi \frac{3}{7})$.

After obtaining $\vh_v^0 = S(\vx_v)$ for all vertices $v$, all models ran 2 rounds of message passing ($T = 2$), except for the GCN and ERGCN which ran 1 round.  Dropout regularization of probability 0.5 was used in all models, applied at the layers specified in the paper that originally introduced the model.  Most models used a hidden state size of $d = 256$, except for a few exceptions where required to fit things onto the GPU.  Full hyperparameter specifications for every model and experiment can be found in the experiment scripts in the code released with the paper.\footnote{\url{https://github.com/mwcvitkovic/Supervised-Learning-on-Relational-Databases-with-Graph-Neural-Networks/tree/master/experiments}}

The readout function $R$ for all GNNs was the Gated Attention Pooling method of \cite{ggnnpaper} followed by a linear transform to obtain logits followed by a softmax layer. The only exception is the PoolMLP model, which uses average pooling of hidden states followed by a 1-hidden-layer MLP as $R$.  The cross entropy loss was used for training, and the AdamW optimizer \citep{adamw} was used to update the model parameters.   All models used early stopping based on the performance on a validation set.

\subsection{Other model implementation}

The Single-table LogReg and MLP models used the same steps (1) and (2) for initializing their input data as the GNN implementation in the previous section.  We did not normalize the inputs to the GBDT model, as the \texttt{LightGBM} library handles this automatically.  The MLP model contained 2 hidden layers, the first with dimension 4x the number of inputs, the second with dimension 2x the number of inputs.  LogReg and MLP were trained with weight decay of 0.01, and MLP also used dropout regularization with probability 0.3.  The cross entropy loss was used for training, and for LogReg and MLP the AdamW optimizer \citep{adamw} was used to update the model parameters.   All models used early stopping based on the performance on a validation set.

DFS features were generated using as many possible aggregation primitives and transformation primitives as offered in the \texttt{featuretools} library, except for the Home Credit Default Risk dataset, which had too many features to make this possible with our hardware.  For that dataset we used the same DFS settings that the \texttt{featuretools} authors did when demonstrating their system on that dataset.\footnote{\url{https://www.kaggle.com/willkoehrsen/home-credit-default-risk-feature-tools}}

\subsection{Hyperparameter optimization and evaluation}

No automatic hyperparameter optimization was used for any models.  We manually looked for reasonable values of the following hyperparameters by comparing, one-at-a-time, their effect on the model's performance on the validation set of the first cross-validation split:
\begin{itemize}
\item Number of leaves (in a GBDT tree)
\item Minimum number of datapoints contained in a leaf (in a GBDT tree)
\item Weight decay
\item Dropout probability
\item Whether to one-hot encode embeddings
\item Whether to oversample the minority class during SGD
\item Readout function
\item Whether to apply Batch Normalization, Layer Normalization, or no normalization
\item Number of layers and message passing rounds
\end{itemize}

Additionally, to find a learning rate for the models trained with SGD, we swept through one epoch of training using a range of learning rates to find a reasonable value, in the style of the FastAI \citep{howard2018fastai} learning rate finder.

Full hyperparameter specifications for every model and experiment can be found in the experiment scripts in the code released with the paper.\footnote{\url{https://github.com/mwcvitkovic/Supervised-Learning-on-Relational-Databases-with-Graph-Neural-Networks/tree/master/experiments}}

Every model was trained and tested on the same five cross-validation splits.  80\% of each split was used for training, and 15\% of that 80\% was used as a validation set.

\section{Dataset Information}\label{sec:suppdatasetinfo}
\subsection{Acquire Valued Shoppers Challenge}

\begin{table}[h]
\caption{Acquire Valued Shoppers Challenge dataset summary information. (\url{www.kaggle.com/c/acquire-valued-shoppers-challenge})}
\label{tab:acquirevaluedinfosupp}
\begin{tabular}{l|l}
\toprule
Train datapoints & 160057 \\
Test datapoints & 151484 \\
Total datapoints & 311541 \\
Dataset size (uncompressed, compressed) & 47 GB, 15 GB \\
\midrule
Node types & 7 \\
Edge types (not including self edges) & 10 \\
Output classes & 2 \\
Class balance in training set & 116619 negative, 43438 positive \\
Types of features in dataset & Categorical, Scalar, Datetime \\
\bottomrule
\end{tabular}
\end{table}

\begin{figure}[h]
  \centering
  \includegraphics[width=0.75\textwidth]{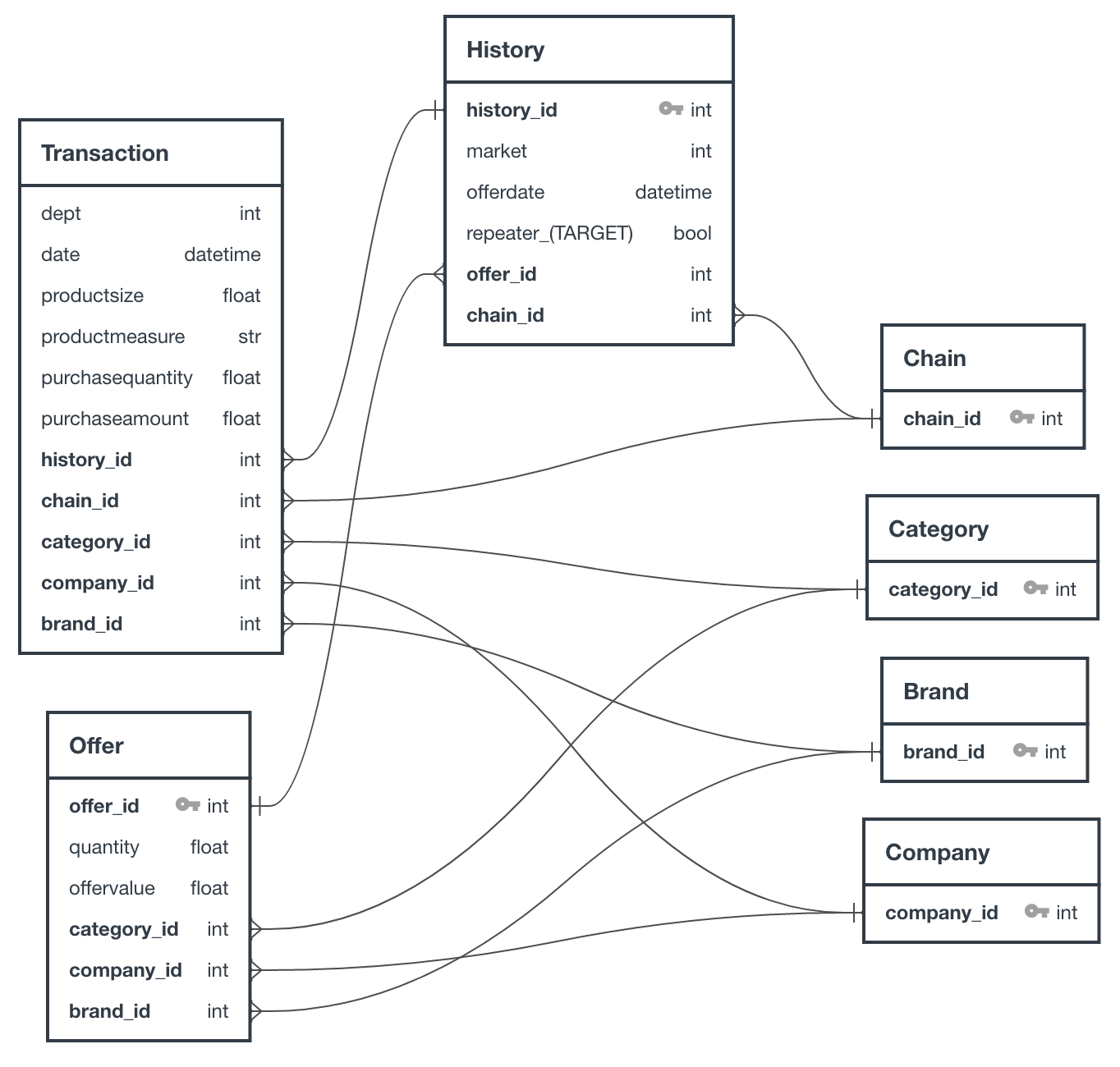}
  \caption{Acquire Valued Shoppers Challenge database schema.}
\label{fig:ValuedShoppersschema}
\end{figure}

\begin{figure}[h!]
  \centering
  \includegraphics[width=\textwidth]{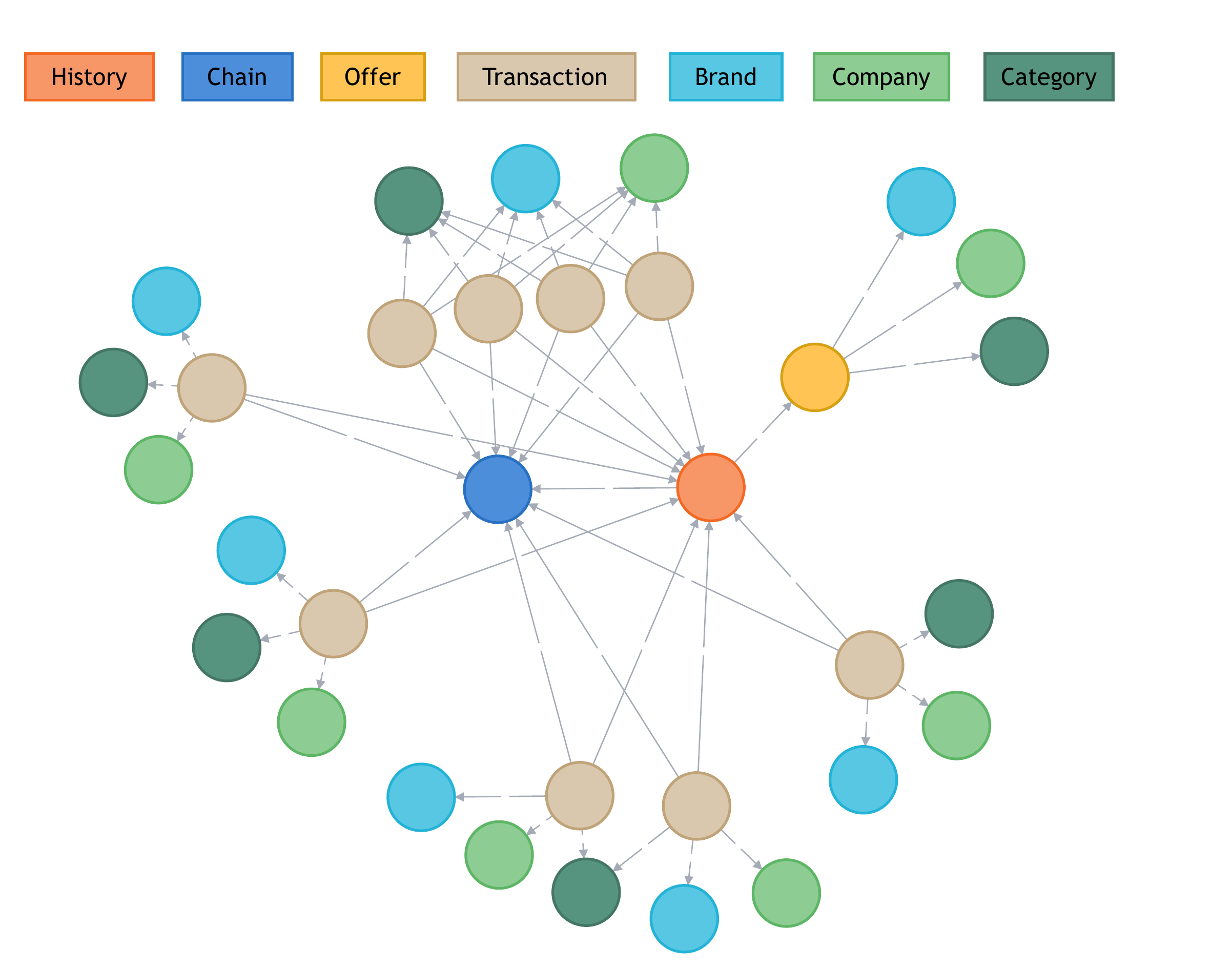}
  \caption{Acquire Valued Shoppers Challenge example datapoint.}
\label{fig:ValuedShoppersexampledp}
\end{figure}

\begin{figure}[h!]
  \centering
  \includegraphics[width=0.65\textwidth]{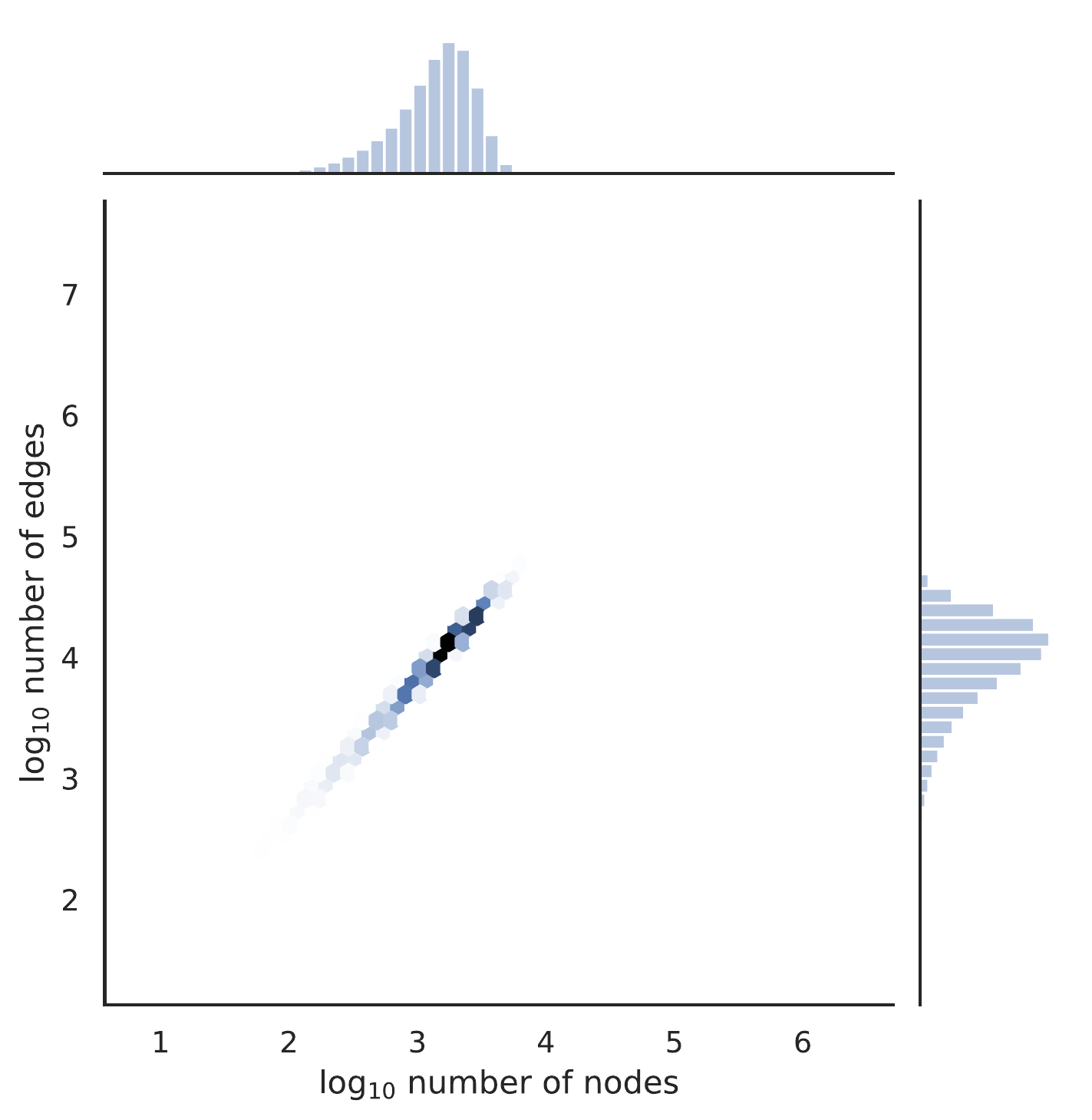}
  \caption{Acquire Valued Shoppers Challenge dataset graph size histogram.  The horizontal axis ranges from the minimum number of nodes in any datapoint to the maximum; likewise with number of edges for the vertical axis.  Thus the distribution of graph sizes in the dataset varies over six orders of magnitude, even if the histogram bins too small to see in the plot.}
\label{fig:Acquiregraphsizehist}
\end{figure}

\newpage

\subsection{Home Credit Default Risk}

\begin{table}[h!]
\caption{Home Credit Default Risk dataset summary information.  (\url{www.kaggle.com/c/home-credit-default-risk/overview/evaluation} ) } 
\label{tab:homecreditinfosupp}
\begin{tabular}{l|p{15em}}
\toprule
Train datapoints & 307511 \\
Test datapoints & 48744 \\
Total datapoints & 356255 \\
Dataset size (uncompressed, compressed) & 6.6 GB, 1.6 GB \\
\midrule
Node types & 7 \\
Edge types (not including self edges) & 9 \\
Output classes & 2 \\
Class balance in training set & 282686 negative, 24825 positive \\
Types of features in dataset & Categorical, Scalar, Geospatial (indirectly), Datetime (indirectly) \\
\bottomrule
\end{tabular}
\end{table}

\begin{figure}[h!]
  \centering
  \includegraphics[width=0.75\textwidth]{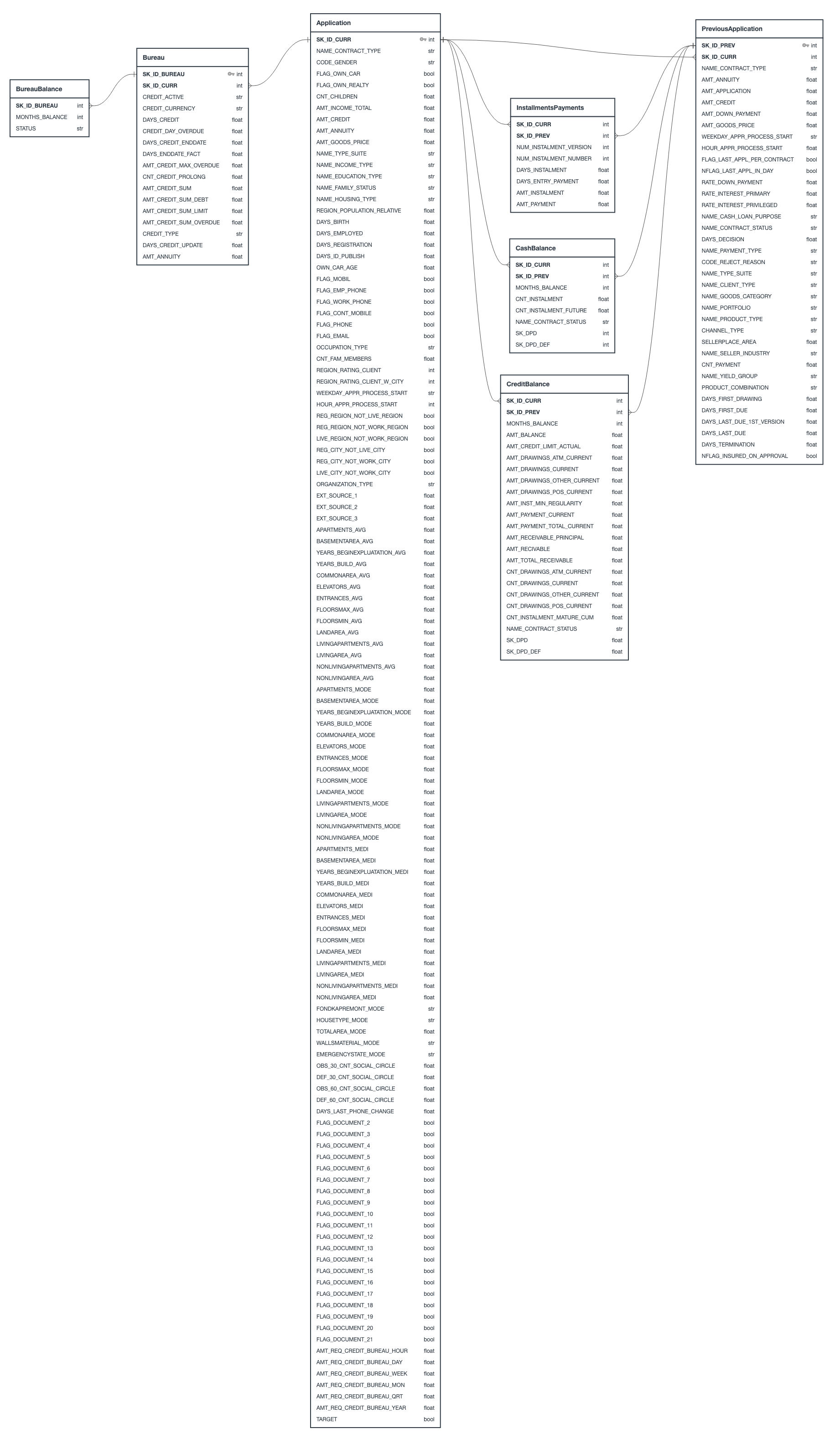}
  \caption{Home Credit Default Risk database schema.}
\label{fig:HomeCreditschema}
\end{figure}

\begin{figure}[h!]
  \centering
  \includegraphics[width=\textwidth]{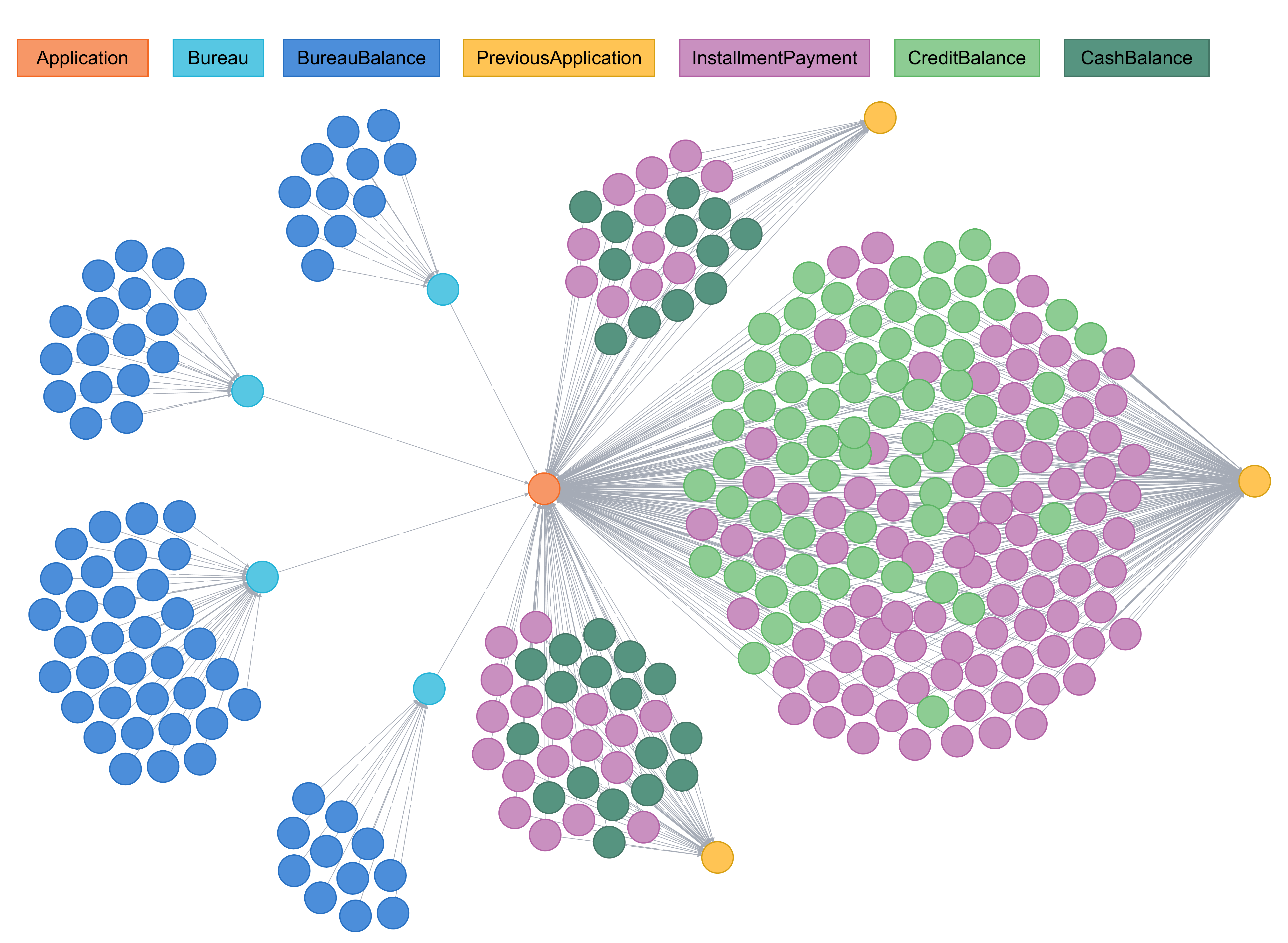}
  \caption{Home Credit Default Risk example datapoint.}
\label{fig:HomeCreditexampledp}
\end{figure}

\begin{figure}[h!]
  \centering
  \includegraphics[width=0.65\textwidth]{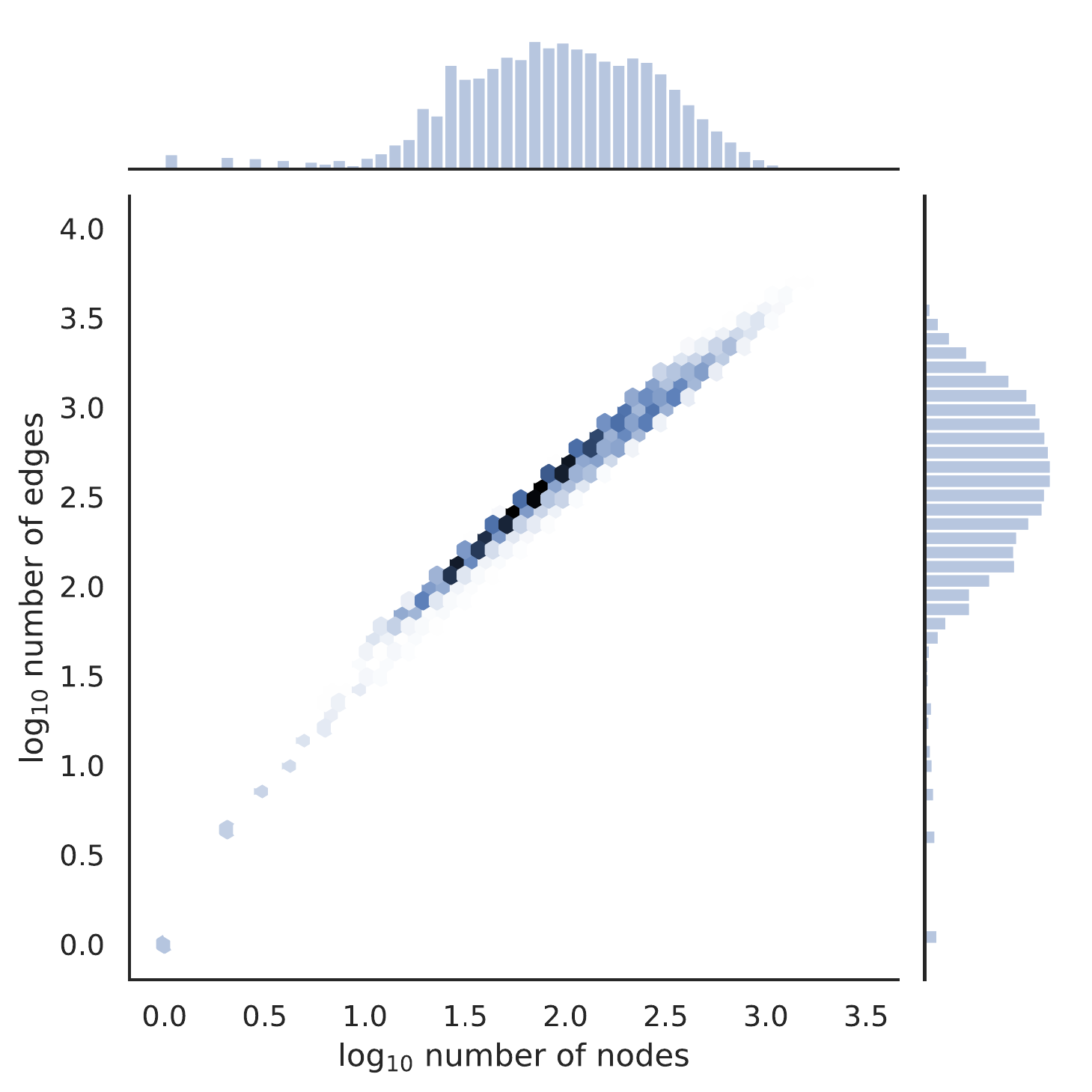}
  \caption{Home Credit Default Risk dataset graph size histogram.  The horizontal axis ranges from the minimum number of nodes in any datapoint to the maximum; likewise with number of edges for the vertical axis.  Thus the distribution of graph sizes in the dataset varies hugely, even if the histogram bins too small to see in the plot.}
\label{fig:Homegraphsizehist}
\end{figure}

\newpage

\subsection{KDD Cup 2014 }

\begin{table}[h!]
\caption{KDD Cup 2014 dataset summary information.   (\url{www.kaggle.com/c/kdd-cup-2014-predicting-excitement-at-donors-choose}) }
\label{tab:kddcupinfosupp}
\begin{tabular}{l|l}
\midrule
Train datapoints & 619326 \\
Test datapoints & 44772 \\
Total datapoints & 664098 \\
Dataset size (uncompressed, compressed) & 3.9 GB, 0.9 GB \\
\midrule
Node types & 4 \\
Edge types (not including self edges) & 4 \\
Output classes & 2 \\
Class balance in training set & 582616 negative, 36710 positive \\
Types of features in dataset & Categorical, Geospatial, Scalar, Textual, Datetime \\
\bottomrule
\end{tabular}
\end{table}

\begin{figure}[h!]
  \centering
  \includegraphics[width=0.75\textwidth]{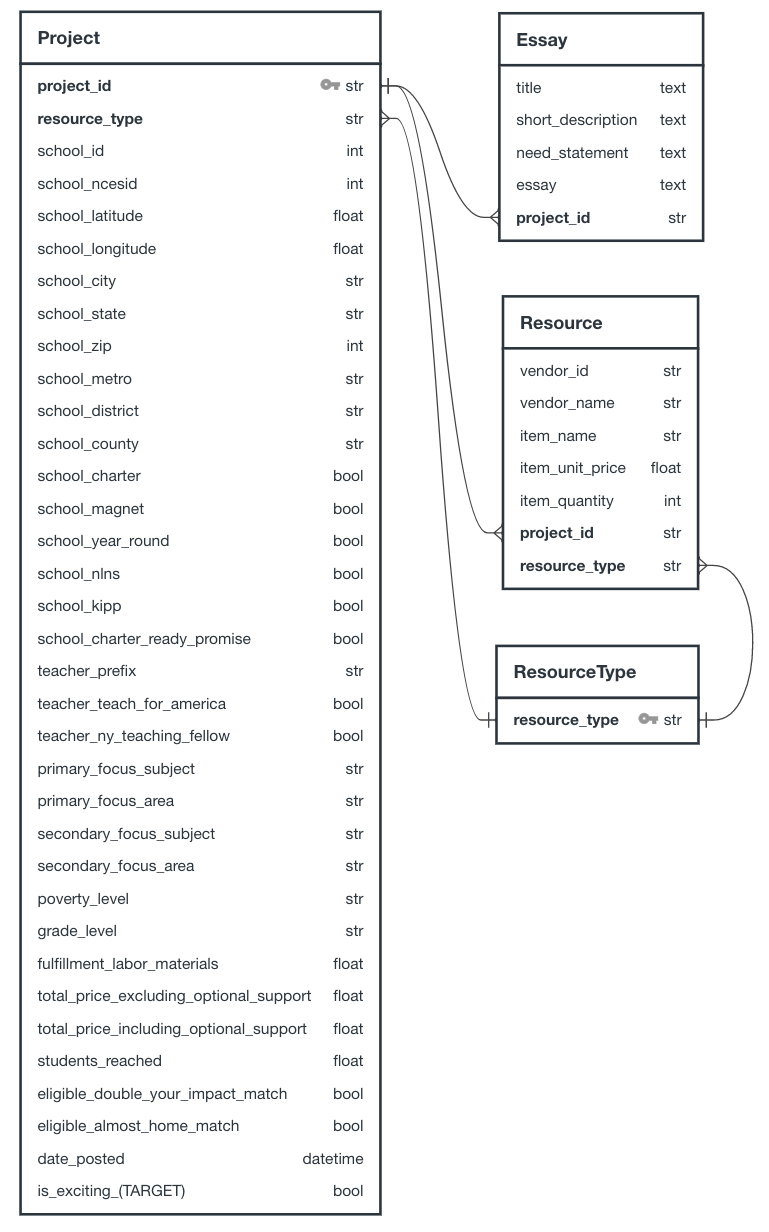}
  \caption{KDD Cup 2014 database schema.}
\label{fig:KDDschema}
\end{figure}

\begin{figure}[h!]
  \centering
  \includegraphics[width=\textwidth]{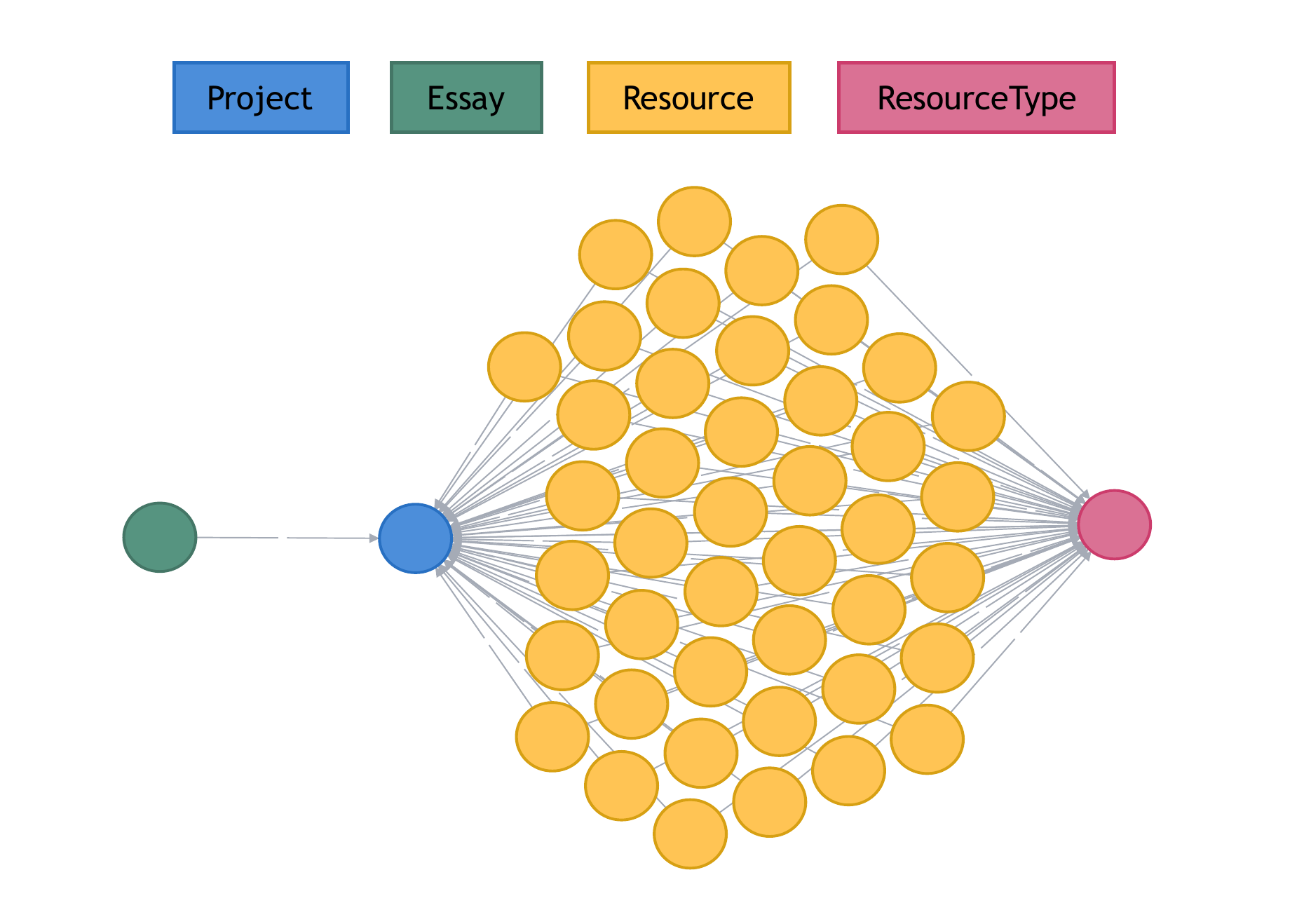}
  \caption{KDD Cup 2014 example datapoint.}
\label{fig:KDDexampledp}
\end{figure}

\begin{figure}[h!]
  \centering
  \includegraphics[width=0.65\textwidth]{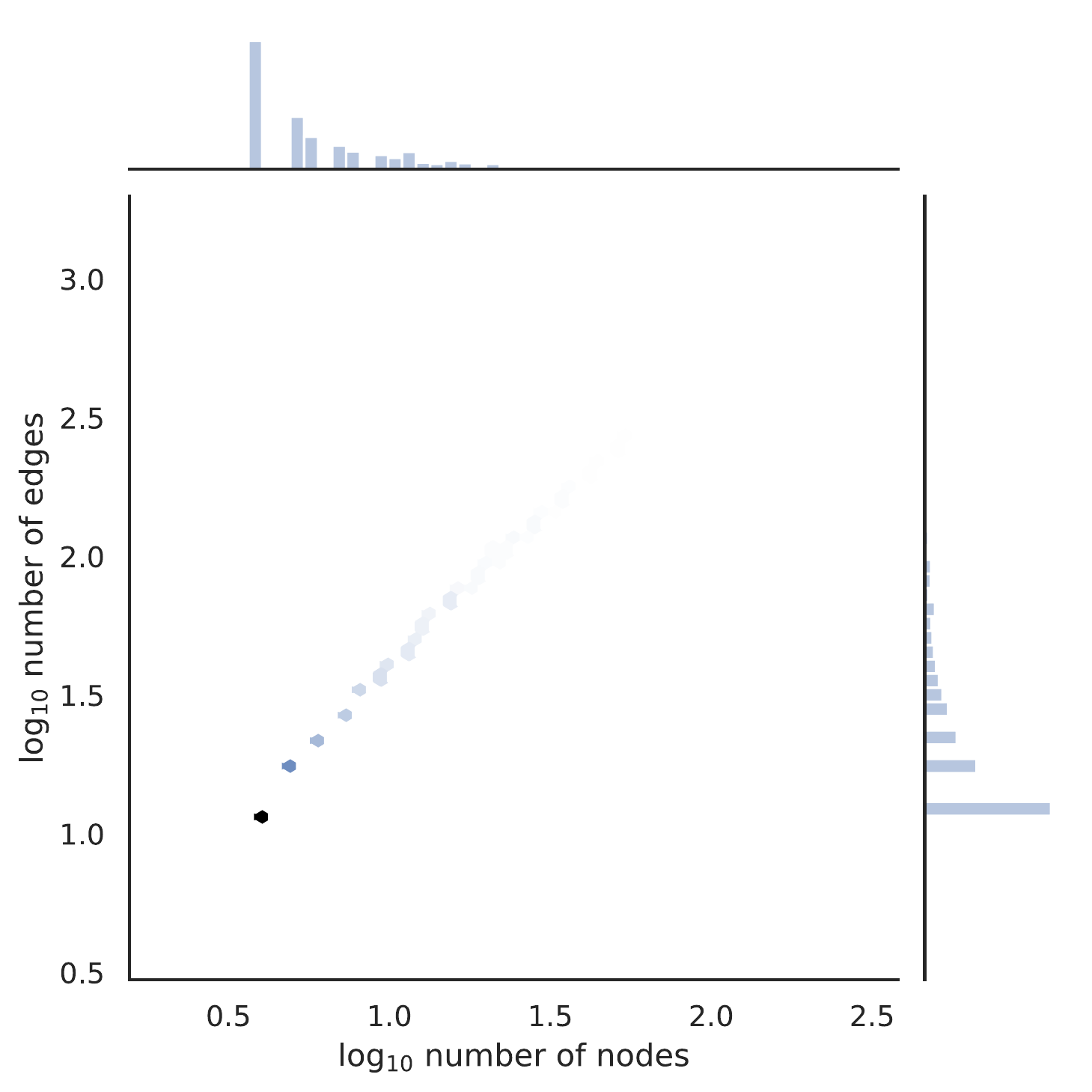}
  \caption{KDD Cup 2014 dataset graph size histogram. The horizontal axis ranges from the minimum number of nodes in any datapoint to the maximum; likewise with number of edges for the vertical axis.  Thus the distribution of graph sizes in the dataset varies hugely, even if the histogram bins too small to see in the plot.}
\label{fig:KDDgraphsizehist}
\end{figure}

\end{document}

%% file: math_commands.tex
\usepackage{amsmath,amsfonts,bm}

\def\eqref#1{equation~\ref{#1}}

\def\1{\bm{1}}

\def\ve{{\bm{e}}}

\def\vh{{\bm{h}}}

\def\vm{{\bm{m}}}

\def\vx{{\bm{x}}}

\def\mT{{\bm{T}}}

\DeclareMathAlphabet{\mathsfit}{\encodingdefault}{\sfdefault}{m}{sl}
\SetMathAlphabet{\mathsfit}{bold}{\encodingdefault}{\sfdefault}{bx}{n}













%% file: neurips_2019.bbl
\begin{thebibliography}{31}
\providecommand{\natexlab}[1]{#1}
\providecommand{\url}[1]{\texttt{#1}}
\expandafter\ifx\csname urlstyle\endcsname\relax
  \providecommand{\doi}[1]{doi: #1}\else
  \providecommand{\doi}{doi: \begingroup \urlstyle{rm}\Url}\fi

\bibitem[Arik and Pfister(2019)]{arik2019tabnet}
Sercan~O Arik and Tomas Pfister.
\newblock Tabnet: Attentive interpretable tabular learning.
\newblock \emph{arXiv preprint arXiv:1908.07442}, 2019.
\newblock URL \url{https://arxiv.org/abs/1908.07442}.

\bibitem[Asay(2016)]{asay_nosql_2016}
Matt Asay.
\newblock {NoSQL} keeps rising, but relational databases still dominate big
  data, April 2016.
\newblock URL
  \url{https://www.techrepublic.com/article/nosql-keeps-rising-but-relational-databases-still-dominate-big-data/}.

\bibitem[Atwood and Towsley(2016)]{atwood2016diffusion}
James Atwood and Don Towsley.
\newblock Diffusion-convolutional neural networks.
\newblock In \emph{Advances in Neural Information Processing Systems}, pages
  1993--2001, 2016.

\bibitem[Battaglia et~al.(2018)Battaglia, Hamrick, Bapst, Sanchez-Gonzalez,
  Zambaldi, Malinowski, Tacchetti, Raposo, Santoro, Faulkner,
  et~al.]{battaglia_relational_2018}
Peter~W Battaglia, Jessica~B Hamrick, Victor Bapst, Alvaro Sanchez-Gonzalez,
  Vinicius Zambaldi, Mateusz Malinowski, Andrea Tacchetti, David Raposo, Adam
  Santoro, Ryan Faulkner, et~al.
\newblock Relational inductive biases, deep learning, and graph networks.
\newblock \emph{arXiv preprint arXiv:1806.01261}, 2018.
\newblock URL \url{https://arxiv.org/abs/1806.01261}.

\bibitem[Getoor and Taskar(2007)]{getoor_introduction_2007}
Lise Getoor and Ben Taskar.
\newblock \emph{Introduction to {Statistical} {Relational} {Learning}}.
\newblock MIT Press, 2007.
\newblock ISBN 978-0-262-07288-5.

\bibitem[Gilmer et~al.(2017)Gilmer, Schoenholz, Riley, Vinyals, and
  Dahl]{gilmer_neural_2017}
Justin Gilmer, Samuel~S. Schoenholz, Patrick~F. Riley, Oriol Vinyals, and
  George~E. Dahl.
\newblock Neural message passing for quantum chemistry.
\newblock In \emph{Proceedings of the 34th International Conference on Machine
  Learning - Volume 70}, ICML'17, pages 1263--1272. JMLR.org, 2017.
\newblock URL \url{http://dl.acm.org/citation.cfm?id=3305381.3305512}.

\bibitem[Guo and Viktor(2008)]{guo_multirelational_2008}
Hongyu Guo and Herna~L Viktor.
\newblock Multirelational classification: a multiple view approach.
\newblock \emph{Knowledge and Information Systems}, 17\penalty0 (3):\penalty0
  287--312, 2008.

\bibitem[Hamilton et~al.(2017)Hamilton, Ying, and
  Leskovec]{hamilton2017inductive}
Will Hamilton, Zhitao Ying, and Jure Leskovec.
\newblock Inductive representation learning on large graphs.
\newblock In \emph{Advances in Neural Information Processing Systems}, pages
  1024--1034, 2017.
\newblock URL \url{https://arxiv.org/abs/1706.02216}.

\bibitem[Howard et~al.(2018)]{howard2018fastai}
Jeremy Howard et~al.
\newblock Fastai.
\newblock \url{https://github.com/fastai/fastai}, 2018.

\bibitem[{International Organization for
  Standardization}(2016)]{international_organization_for_standardization_iso/iec_2016}
{International Organization for Standardization}.
\newblock {ISO}/{IEC} 9075-1:2016: {Information} technology -- {Database}
  languages -- {SQL} -- {Part} 1: {Framework} ({SQL}/{Framework}), December
  2016.
\newblock URL
  \url{http://www.iso.org/cms/render/live/en/sites/isoorg/contents/data/standard/06/35/63555.html}.

\bibitem[{Kaggle, Inc.}(2017)]{kaggle_inc_state_2017}
{Kaggle, Inc.}
\newblock The {State} of {ML} and {Data} {Science} 2017, 2017.
\newblock URL \url{https://www.kaggle.com/surveys/2017}.

\bibitem[Kanter and Veeramachaneni(2015)]{kanter_deep_2015}
James~Max Kanter and Kalyan Veeramachaneni.
\newblock Deep feature synthesis: {Towards} automating data science endeavors.
\newblock In \emph{2015 {IEEE} {International} {Conference} on {Data} {Science}
  and {Advanced} {Analytics} ({DSAA})}, pages 1--10, October 2015.
\newblock \doi{10.1109/DSAA.2015.7344858}.
\newblock URL \url{https://www.jmaxkanter.com/static/papers/DSAA_DSM_2015.pdf}.

\bibitem[Ke et~al.(2017)Ke, Meng, Finley, Wang, Chen, Ma, Ye, and
  Liu]{ke2017lightgbm}
Guolin Ke, Qi~Meng, Thomas Finley, Taifeng Wang, Wei Chen, Weidong Ma, Qiwei
  Ye, and Tie-Yan Liu.
\newblock Lightgbm: A highly efficient gradient boosting decision tree.
\newblock In \emph{Advances in Neural Information Processing Systems}, pages
  3146--3154, 2017.
\newblock URL
  \url{https://papers.nips.cc/paper/6907-lightgbm-a-highly-efficient-gradient-boosting-decision-tree.pdf}.

\bibitem[Kipf and Welling(2016)]{kipf2016semi}
Thomas~N Kipf and Max Welling.
\newblock Semi-supervised classification with graph convolutional networks.
\newblock In \emph{International Conference on Learning Representations}, 2016.
\newblock URL \url{https://arxiv.org/abs/1609.02907}.

\bibitem[Koller et~al.(2007)Koller, Friedman, D{\v{z}}eroski, Sutton, McCallum,
  Pfeffer, Abbeel, Wong, Heckerman, Meek, et~al.]{koller2007introduction}
Daphne Koller, Nir Friedman, Sa{\v{s}}o D{\v{z}}eroski, Charles Sutton, Andrew
  McCallum, Avi Pfeffer, Pieter Abbeel, Ming-Fai Wong, David Heckerman, Chris
  Meek, et~al.
\newblock \emph{Introduction to statistical relational learning}, chapter~2,
  pages 129--174.
\newblock MIT press, 2007.

\bibitem[Lam et~al.(2017)Lam, Thiebaut, Sinn, Chen, Mai, and
  Alkan]{lam_one_2017}
Hoang~Thanh Lam, Johann-Michael Thiebaut, Mathieu Sinn, Bei Chen, Tiep Mai, and
  Oznur Alkan.
\newblock One button machine for automating feature engineering in relational
  databases.
\newblock \emph{arXiv:1706.00327 [cs]}, June 2017.
\newblock URL \url{http://arxiv.org/abs/1706.00327}.
\newblock arXiv: 1706.00327.

\bibitem[Lam et~al.(2018)Lam, Minh, Sinn, Buesser, and
  Wistuba]{lam_neural_2018}
Hoang~Thanh Lam, Tran~Ngoc Minh, Mathieu Sinn, Beat Buesser, and Martin
  Wistuba.
\newblock Neural {Feature} {Learning} {From} {Relational} {Database}.
\newblock \emph{arXiv:1801.05372 [cs]}, January 2018.
\newblock URL \url{http://arxiv.org/abs/1801.05372}.
\newblock arXiv: 1801.05372.

\bibitem[Li et~al.(2016)Li, Tarlow, Brockschmidt, and Zemel]{ggnnpaper}
Yujia Li, Daniel Tarlow, Marc Brockschmidt, and Richard Zemel.
\newblock Gated graph sequence neural networks.
\newblock In \emph{International Conference on Learning Representations}, 2016.
\newblock URL \url{https://arxiv.org/abs/1511.05493}.

\bibitem[Loshchilov and Hutter(2017)]{adamw}
Ilya Loshchilov and Frank Hutter.
\newblock Decoupled weight decay regularization.
\newblock In \emph{International Conference on Learning Representations}, 2017.
\newblock URL \url{https://arxiv.org/abs/1711.05101}.

\bibitem[Luzhnica et~al.(2019)Luzhnica, Day, and Li{\`o}]{Luzhnica2019OnGC}
Enxhell Luzhnica, Ben Day, and Pietro Li{\`o}.
\newblock On graph classification networks, datasets and baselines.
\newblock \emph{ArXiv}, abs/1905.04682, 2019.
\newblock URL \url{https://arxiv.org/abs/1905.04682}.

\bibitem[Murphy et~al.(2019)Murphy, Srinivasan, Rao, and
  Ribeiro]{murphy2019relational}
Ryan Murphy, Balasubramaniam Srinivasan, Vinayak Rao, and Bruno Ribeiro.
\newblock Relational pooling for graph representations.
\newblock In Kamalika Chaudhuri and Ruslan Salakhutdinov, editors,
  \emph{Proceedings of the 36th International Conference on Machine Learning},
  volume~97 of \emph{Proceedings of Machine Learning Research}, pages
  4663--4673, Long Beach, California, USA, 09--15 Jun 2019. PMLR.
\newblock URL \url{http://proceedings.mlr.press/v97/murphy19a.html}.

\bibitem[Paszke et~al.(2019)Paszke, Gross, Massa, Lerer, Bradbury, Chanan,
  Killeen, Lin, Gimelshein, Antiga, Desmaison, Kopf, Yang, DeVito, Raison,
  Tejani, Chilamkurthy, Steiner, Fang, Bai, and Chintala]{pytorchpaper}
Adam Paszke, Sam Gross, Francisco Massa, Adam Lerer, James Bradbury, Gregory
  Chanan, Trevor Killeen, Zeming Lin, Natalia Gimelshein, Luca Antiga, Alban
  Desmaison, Andreas Kopf, Edward Yang, Zachary DeVito, Martin Raison, Alykhan
  Tejani, Sasank Chilamkurthy, Benoit Steiner, Lu~Fang, Junjie Bai, and Soumith
  Chintala.
\newblock Pytorch: An imperative style, high-performance deep learning library.
\newblock In H.~Wallach, H.~Larochelle, A.~Beygelzimer, F.~d'Alch\'{e} Buc,
  E.~Fox, and R.~Garnett, editors, \emph{Advances in Neural Information
  Processing Systems 32}, pages 8024--8035. Curran Associates, Inc., 2019.
\newblock URL
  \url{http://papers.neurips.cc/paper/9015-pytorch-an-imperative-style-high-performance-deep-learning-library.pdf}.

\bibitem[Schlichtkrull et~al.(2018)Schlichtkrull, Kipf, Bloem, Van Den~Berg,
  Titov, and Welling]{Schlichtkrull2017ModelingRD}
Michael Schlichtkrull, Thomas~N Kipf, Peter Bloem, Rianne Van Den~Berg, Ivan
  Titov, and Max Welling.
\newblock Modeling relational data with graph convolutional networks.
\newblock In \emph{European Semantic Web Conference}, pages 593--607. Springer,
  2018.
\newblock URL \url{https://arxiv.org/abs/1703.06103}.

\bibitem[Shi et~al.(2016)Shi, Li, Zhang, Sun, and Philip]{shi2016survey}
Chuan Shi, Yitong Li, Jiawei Zhang, Yizhou Sun, and S~Yu Philip.
\newblock A survey of heterogeneous information network analysis.
\newblock \emph{IEEE Transactions on Knowledge and Data Engineering},
  29\penalty0 (1):\penalty0 17--37, 2016.
\newblock URL \url{https://arxiv.org/pdf/1511.04854.pdf}.

\bibitem[{Stack Exchange, Inc.}(2018)]{stack_exchange_inc_stack_2018}
{Stack Exchange, Inc.}
\newblock Stack {Overflow} {Developer} {Survey} 2018, 2018.
\newblock URL \url{https://insights.stackoverflow.com/survey/2018/}.

\bibitem[Veličković et~al.(2018)Veličković, Cucurull, Casanova, Romero,
  Lio, and Bengio]{gatpaper}
Petar Veličković, Guillem Cucurull, Arantxa Casanova, Adriana Romero, Pietro
  Lio, and Yoshua Bengio.
\newblock Graph attention networks.
\newblock In \emph{International Conference on Learning Representations}, 2018.
\newblock URL \url{https://openreview.net/forum?id=rJXMpikCZ}.

\bibitem[Wang et~al.(2019{\natexlab{a}})Wang, Yu, Zheng, Gan, Gai, Ye, Li,
  Zhou, Huang, Ma, Huang, Guo, Zhang, Lin, Zhao, Li, Smola, and
  Zhang]{dglpaper}
Minjie Wang, Lingfan Yu, Da~Zheng, Quan Gan, Yu~Gai, Zihao Ye, Mufei Li,
  Jinjing Zhou, Qi~Huang, Chao Ma, Ziyue Huang, Qipeng Guo, Hao Zhang, Haibin
  Lin, Junbo Zhao, Jinyang Li, Alexander~J Smola, and Zheng Zhang.
\newblock Deep graph library: Towards efficient and scalable deep learning on
  graphs.
\newblock \emph{ICLR Workshop on Representation Learning on Graphs and
  Manifolds}, 2019{\natexlab{a}}.
\newblock URL \url{https://arxiv.org/abs/1909.01315}.

\bibitem[Wang et~al.(2019{\natexlab{b}})Wang, Ji, Shi, Wang, Ye, Cui, and
  Yu]{wang2019heterogeneous}
Xiao Wang, Houye Ji, Chuan Shi, Bai Wang, Yanfang Ye, Peng Cui, and Philip~S
  Yu.
\newblock Heterogeneous graph attention network.
\newblock In \emph{The World Wide Web Conference}, pages 2022--2032. ACM,
  2019{\natexlab{b}}.
\newblock URL \url{https://arxiv.org/abs/1903.07293}.

\bibitem[Wolpert(1992)]{wolpert1992stacked}
David~H Wolpert.
\newblock Stacked generalization.
\newblock \emph{Neural networks}, 5\penalty0 (2):\penalty0 241--259, 1992.

\bibitem[Wu et~al.(2019)Wu, Pan, Chen, Long, Zhang, and
  Yu]{wu_comprehensive_2019}
Zonghan Wu, Shirui Pan, Fengwen Chen, Guodong Long, Chengqi Zhang, and Philip~S
  Yu.
\newblock A comprehensive survey on graph neural networks.
\newblock \emph{arXiv preprint arXiv:1901.00596}, 2019.
\newblock URL \url{https://arxiv.org/abs/1901.00596}.

\bibitem[Xu et~al.(2019)Xu, Hu, Leskovec, and Jegelka]{ginpaper}
Keyulu Xu, Weihua Hu, Jure Leskovec, and Stefanie Jegelka.
\newblock How powerful are graph neural networks?
\newblock In \emph{International Conference on Learning Representations}, 2019.
\newblock URL \url{https://openreview.net/forum?id=ryGs6iA5Km}.

\end{thebibliography}
